\documentclass[a4paper,11pt]{article}

\usepackage{breakurl}
\usepackage[breaklinks=true]{hyperref}
\usepackage{breakcites}

\usepackage{amssymb,amsmath,array,bm}
\usepackage{booktabs}

\usepackage{tikz}
\usetikzlibrary{arrows}

\usepackage{pgfplots}
\pgfplotsset{compat=1.15}
\usetikzlibrary{arrows}

\usepackage[a4paper, left=3cm, right=2.5cm, top=3cm, bottom=3cm]{geometry}

\usepackage{authblk}
\usepackage{doi}

\usepackage{amsthm}
\newtheorem{theorem}{Theorem}
\newtheorem{definition}{Definition}
\newcommand{\R}{\mathbb{R}}
\newcommand{\N}{\mathbb{N}}

\usepackage{tikz}
\usetikzlibrary{arrows.meta, shapes.geometric, positioning, fit, backgrounds, calc}
\usepackage{tcolorbox}
\usepackage{xcolor}
\usepackage{booktabs}      
\usepackage{colortbl}      
\usepackage{array}         
\usepackage{subcaption}
\usepackage{csquotes}
\tcbuselibrary{skins, breakable}
\definecolor{okgreen1}{RGB}{0, 128, 0}
\definecolor{okorange}{RGB}{230, 100, 0}
\usetikzlibrary{arrows.meta, positioning, calc}

\definecolor{uniblue}{RGB}{0,120,180}
\definecolor{headblue}{RGB}{0,90,140}
\definecolor{okgreen}{RGB}{40,150,80}
\definecolor{warnorange}{RGB}{220,100,30}
\definecolor{lightblue}{RGB}{210,235,250}
\definecolor{unigrey}{RGB}{100,100,100}
\usepackage[utf8]{inputenc}
\usepackage[T1]{fontenc}
\usepackage[english]{babel}
\usepackage{csquotes}

\title{Robust Deep Prototype Classifiers via Lipschitz Training}

\author[1]{Adia Khalid Khan}
\author[1]{Lukas Gehring}
\author[1]{Benjamin Paaßen}
\affil[1]{Faculty of Technology, Bielefeld University}

\date{11-06-2026}

\begin{document}

\maketitle

\pagestyle{myheadings}
\markright{Robustness via Lipschitz Training}

\begin{abstract}
Robustness and interpretability are interrelated and desirable properties of machine learning models. A model that changes its prediction under imperceptible perturbations behaves inconsistently with human-recognizable features and, hence, is not interpretable. Prototype-based classifiers such as Generalized Learning Vector Quantization (GLVQ) address both at once. They are interpretable by construction because they make the decision by comparing the input to the prototypical example; and their hypothesis margin yields a closed form robustness certificate.
However, on high-dimensional data, such as images, these shallow models need deep embedding networks to achieve high accuracy, and this voids the robustness certificate. The margin is a statement about distances in the embedding space, and an unconstrained network can map an arbitrarily small input perturbation to arbitrarily large attacks in the embedding space. We close this gap with a composition theorem: if the embedding network is locally $L$-Lipschitz around a point, and the classifier is $\epsilon$-robust around its embedding, then the full model is $\epsilon/L$-robust in the input space. 
Based on this theorem, we propose a novel Lipschitz training scheme to achieve $L$-Lipschitz properties for embedding networks. In experiments, we evaluate three classification heads (linear, linear Support Vector Machine (SVM), and GLVQ) crossed with two embedding networks (standard and Lipschitz-constrained) on MNIST, CIFAR-10, and Imagenette-160, under Fast Gradient Sign (FGSM), Projected Gradient Descent (PGD), and Carlini-Wagner (C\&W) attacks. The empirical evaluation shows that combining a prototype-based head with a Lipschitz-constrained embedding provides the strongest robustness. For example, on MNIST, the combined model retains roughly $52.9\%$ accuracy under PGD at $\epsilon = 0.20$, while all unconstrained models essentially collapse. This suggests that the Lipschitz constraint provides an additional robustness benefit in addition to the advantage provided by the prototype-based head. However, since the Lipschitz constraint is enforced by sampling rather than by construction, the reported robustness is empirical rather than certified.
\end{abstract}

\section{Introduction}\label{sec:intro}
Machine learning systems have achieved remarkable breakthroughs in recent years, enabling applications across a wide range of domains, including medical imaging, computer vision, autonomous driving, and security monitoring. In these safety-critical applications, erroneous predictions or decisions can have potentially catastrophic consequences~\cite{Zhou2022}. Therefore, such applications demand not only accurate predictions but robustness and interpretability~\cite{carlini2019evaluating, Goepfert2020}; robust in the sense that decisions should remain stable under small changes in the input that a human would consider insignificant~\cite{szegedy2013intriguing, goodfellow2014explaining, xu2020adversarial}; and interpretable in the sense that users should be able to understand why a particular decision was made~\cite{gunning2850darpa, xu2019explainable, biehl2016wires}, to assess the system before deployment and to diagnose potential failures afterwards. High accuracy on a test set alone does not guarantee either of these properties.

Szegedy et al.~\cite{szegedy2013intriguing} were among the first to demonstrate this gap. They showed that small, often imperceptible changes to the input can completely change a network's prediction, and that such changes often transfer between independently trained models. Over the past decade, researchers have confirmed that even models with excellent accuracy on clean data can be reduced to near-zero accuracy under white-box attacks~\cite{goodfellow2014explaining, madry2017towards, carlini2017towards}. 
Vulnerability to adversarial attacks can also indicate a lack of interpretability, as the model may rely on features that are not meaningful or recognizable to humans. Therefore, robustness and interpretability are closely connected rather than being distinct goals. Both assess whether the model makes decisions based on features that humans can understand and recognize. This connection is the main focus of this paper.

Two main defense approaches (adversarial training and Lipschitz-based methods) have been proposed in the literature, each with its own strengths and drawbacks. Adversarial training adds adversarial examples into the training data and remains the most effective empirical defense~\cite{madry2017towards, zhang2019trades}. However, this approach requires an inner optimization at every training step, trades clean accuracy for robustness~\cite{tsipras2019robustness, zhang2019trades}, and provides no formal guarantees: robustness is only certified against the attacks that were actually tested. Lipschitz-based methods, on the other hand, bound how much the model can amplify input differences, providing an attack-independent guarantee~\cite{fazlyab2019efficient, scaman2018lipschitz, hoffman2019robust}. Yet a Lipschitz bound alone is not a defense. In fact, the most Lipschitz-stable model is the constant function, which is useless for prediction~\cite{zuhlke2025adversarial}. 
Therefore, we propose to combine a Lipschitz constraint for a deep embedding network with the robustness guarantees of shallow classification models.

Prototype-based classifiers are such intrinsically robust classification models---and they additionally offer interpretability as a bonus. A nearest prototype classifier assigns an input to the class of its closest prototype. Thus, its decisions are interpretable as long as the prototypes themselves are interpretable~\cite{biehl2016wires, davoodi2023}. Since the decision depends solely on distances to a fixed set of prototypes, the distance to the decision boundary can be calculated in closed form. This distance, known as the \emph{hypothesis margin}~\cite{crammer2002}, was originally introduced to derive generalization bounds for Learning Vector Quantization (LVQ)---a specific family of prototype-based classifiers~\cite{crammer2002, hammer2005}. Saralajew et al.~\cite{saralajew2020npc} later showed that it also yields an adversarial robustness guarantee. 
Subsequent work has tightened these bounds further and has shown that prototype classifiers can be competitive with certified neural networks~\cite{Voracek2022}, while empirical studies report that GLVQ models are robust even before any explicit certification~\cite{SaralajewGLVQ}.

The challenge, however, is that LVQ operates on distances in the input space, where the Euclidean distance is not meaningful for image data~\cite{Khalid2025ContEx}. The common solution is a hybrid approach: train a deep network $\phi$ that embeds inputs nonlinearly, and then use prototype-based decisions in this embedding space~\cite{chen2019protopnet, li2018deep, snell2017prototypical}. While this approach improves accuracy, it implicitly compromises the formal robustness guarantee. The certificate of a nearest prototype classifier is only valid in the space where the prototypes live. However, an unconstrained embedding network can map an arbitrarily small input change to an arbitrarily large change in the embedding space~\cite{Khalid2026}. Therefore, a certified radius in the embedding space does not provide information regarding robustness in the original input space---unless we have a guarantee that distances in the embedding space are bounded by distances in the input space, which is precisely what a Lipschitz constraint does.
As illustrated in Figure~\ref{fig:mainidea} (compare the top-left and top-right panels), without a Lipschitz constraint, small changes in the input space can result in large changes in the embedding space, invalidating the robustness guarantee.

Our approach is a combination of prototype-based classification and Lipschitz bounds to achieve better adversarial robustness for deep prototype classifiers. In more detail, our main contributions are:
\begin{itemize}
    \item A composition theorem (Theorem~\ref{thm:main}) that transfers local $\epsilon$-robustness guarantees in the embedding space to the input space using $L$-Lipschitz continuity of the deep embedding network, yielding a radius of $\epsilon/L$ around each correctly classified point. 
	
   \item A Lipschitz loss and corresponding training scheme which enhances the Lipschitz continuity of deep embedding networks by sampling Gaussian noise around the training data during training and punishes violations of the desired Lipschitz constant $L$. Because sampling Gaussian noise is cheap, this is substantially cheaper than adversarial training~\cite{madry2017towards} and less restrictive than a global Lipschitz bound.
	
   \item An evaluation on three data sets (MNIST, CIFAR-10, and Imagenette-160) using three classifiers (linear, linear SVM, and GLVQ) and two embedding networks (standard and Lipschitz-constrained) under three adversarial attacks (FGSM, PGD, and C\&W),
\end{itemize}

This paper is an extension of the conference contribution ESANN~\cite{Khalid2026}. The present paper extends it in four ways: evaluation on three benchmarks instead of one, a sensitivity analysis over the Lipschitz training hyperparameters, the addition of the Carlini--Wagner (C\&W) attacks and Projected Gradient Descent (PGD) attacks, and a measurement of residual constraint-violation rates that qualifies the transferred radius as empirical.

\section{Background and Related Work}\label{sec:background_Study}
This section reviews the prior work on which our contribution builds, focusing on two properties of machine learning models: \emph{adversarial robustness} -- the stability of a model's prediction under small input perturbations ~\cite{szegedy2013intriguing} -- and \emph{interpretability} -- the extent to which a human can understand the model's decision process~\cite{biran2017explanation}.

Adversarial perturbations, defined as imperceptible changes to the input that flip a model's prediction, reveal both a security flaw and a fundamental failure of interpretability, as the model's internal representations diverge from human perception~\cite{xu2020adversarial, Zhou2022, Goepfert2020}. Even systems that achieve state-of-the-art accuracy can be successfully attacked, suffering substantial drops in performance~\cite{carlini2019evaluating, Ilievski2025}. This phenomenon was first systematically studied by Szegedy et al., who demonstrated that small, often imperceptible perturbations can cause misclassification and may transfer between independently trained models~\cite{szegedy2013intriguing}. Formally, given a classifier $f$ with parameters $\theta$, an input-label pair $(x, y)$, and a loss function $\mathcal{L}$, an adversarial example is found by solving the inner maximization problem~\cite{madry2017towards, szegedy2013intriguing}:
        \begin{equation}
        \max_{\eta : \|\eta\|_p \le \epsilon} \mathcal{L}(\theta, x + \eta, y)
        \end{equation}
where $\eta$ is an adversarial perturbation, constrained to an $\ell_p$-norm ball of radius $\epsilon$~\cite{goodfellow2014explaining, madry2017towards}. Such perturbations are commonly constrained using $\ell_\infty$, $\ell_2$, or $\ell_0$ norms~\cite{carlini2017towards}. Based on the level of access to the target model, attacks are typically categorized as white-box, where the adversary has full access to the model parameters and gradients, or black-box, where attacks rely on transferability without direct access~\cite{yuan2019adversarial, chakraborty2021survey, papernot2017practical}. Since white-box attacks provide direct access to the model's vulnerabilities~\cite{yuan2019adversarial}, they represent the most challenging threat scenario and are the focus of rigorous robustness evaluation.

Among white-box attacks, the fast gradient sign method (FGSM(~\cite{goodfellow2014explaining} is a single-step $\ell_\infty$ attack that perturbs the input by $\epsilon$ in the direction of the sign of the loss gradient. It is computationally efficient and serves as a lower bound on attack strength; a model that fails against FGSM cannot be considered robust in any meaningful sense. Projected gradient descent (PGD)~\cite{madry2017towards} extends this to a multi-step first-order adversary and is substantially stronger than single-step variants~\cite{madry2017towards, dong2018boosting, shafahi2019adversarial}. Madry et al.~\cite{madry2017towards} argue that PGD with random restarts approximates the strongest first-order adversary, establishing it as the standard reference attack for $\ell_\infty$ robustness. Beyond FGSM and PGD, 
the Carlini \& Wagner (C\&W) attack~\cite{carlini2017towards} formulates adversarial example generation as an optimization problem that jointly minimizes the perturbation norm and a surrogate of the classification margin under $\ell_0$, $\ell_2$, or $\ell_\infty$ constraints. This attack achieves 100\% success on both distilled and undistilled networks, revealing that defensive distillation, previously claimed to reduce attack success to 0.5\%, provides little real security benefit~\cite{carlini2017towards}. Other notable attacks include DeepFool~\cite{moosavi2016deepfool}, which searches for the minimal perturbation crossing the decision boundary by iteratively linearizing it; saliency-guided attacks such as JSMA, which target $\ell_0$ sparsity~\cite{papernot2016limitations}; and AutoAttack~\cite{croce2020reliable}, a parameter-free ensemble of four complementary attacks designed for reliable robustness evaluation. Croce and Hein~\cite{croce2020reliable} showed that published defenses frequently report inflated robustness: across 49 deterministic defenses, AutoAttack obtained lower robust accuracy than originally reported in all but one case, with drops exceeding 10 percentage points in 13 cases and by more than 30 percentage points in eight~\cite{croce2020reliable}.

The design of effective defense mechanisms against adversarial perturbations must navigate a trade-off: improving robustness against adversarial perturbations often reduces accuracy on clean data~\cite{zhang2019trades}. Moreover, training robust classifiers is computationally expensive and does not provide formal guarantees. 
Despite these costs, robustly trained models learn feature representations that are more closely aligned with human-perceptible features~\cite{tsipras2019robustness}, suggesting that robustness and interpretability are interrelated.

To enhance adversarial robustness, numerous defense strategies have been proposed, ranging from adversarial training~\cite{goodfellow2014explaining, madry2017towards} to regularization techniques~\cite{ross2018improving}. Adversarial training, which augments the training set with adversarial examples, has emerged as one of the most effective empirical defenses. Zhang et al.~\cite{zhang2019trades} formalized the robustness--accuracy trade-off through TRADES, which achieves a theoretically principled balance between standard accuracy and adversarial robustness, winning first place in the NeurIPS 2018 Adversarial Vision Challenge and surpassing the runner-up by 11.41\% in mean $\ell_2$ perturbation distance. However, adversarial training is computationally expensive and does not provide formal guarantees.

Lipschitz theory instead offers guarantees that hold independently of any particular attack~\cite{zuhlke2025adversarial}: the Lipschitz constant $L$ upper-bounds the amplification of input differences and is thus a direct measure of perturbation sensitivity~\cite{fazlyab2019efficient, scaman2018lipschitz}. This motivates constraining $L$ during training via Jacobian or spectral norm regularization~\cite{zhao2021ernn, yoshida2017spectral}, input-gradient regularization~\cite{finlay2021scaleable}, or weight-matrix conditioning~\cite{feng2025lipschitz}. For instance, Hoffman et al.~\cite{hoffman2019robust} approximate local Lipschitz constraints by penalizing the Jacobian norm via random projection (a computationally efficient estimator of local sensitivity), without generating adversarial perturbations. The framework has two well-known limitations. First, the most Lipschitz-stable model is the constant function with $L = 0$, which is useless for prediction~\cite{zuhlke2025adversarial}.
Second, computing $L$ exactly is NP-hard even for two-layer networks~\cite{scaman2018lipschitz}; consequently, all practical estimators must trade tightness against scalability~\cite{fazlyab2019efficient}. 
Furthermore, Ross and Doshi-Velez~\cite{ross2018improving} found that penalizing input-gradient norms improves robustness to transferred adversarial examples while simultaneously making input gradients more interpretable as explanations. Specifically, human evaluators rated the resulting misclassifications as more plausible than those of other defended models, suggesting a strong fundamental connection between robustness and interpretability \cite{vadillo2025adversarial}. 
Some intrinsically interpretable models also provide robustness guarantees. Prototype-based models perform classification by comparing input data points to prototypical examples for each class, which makes them interpretable by construction: as long as the prototypes themselves are interpretable, so are the decisions~\cite{biehl2016wires, davoodi2023}. Formally, a nearest prototype classifier assigns an input data point $x$ to the label of the closest prototype $w$ from a learned, finite set of prototypes $W \subset \mathcal{X}$ under a dissimilarity $d : \mathcal{X} \times \mathcal{X} \to \mathbb{R}_{\geq 0}$~\cite{saralajew2020npc}. Learning Vector Quantization (LVQ)~\cite{kohonen1990} is the foundational nearest prototype classifier; Generalized LVQ (GLVQ)~\cite{sato1995generalized} replaces the heuristic update with a differentiable cost function, and further extensions either learn the metric itself~\cite{hammer2002grlvq, schneider2009gmlvq} or introduce probabilistic variants~\cite{seo2003}. The connection between prototype models and robustness is structural rather than incidental. Crammer et al.~\cite{crammer2002} showed that LVQ-type learning maximizes a \emph{hypothesis margin}, and Hammer et al.~\cite{hammer2005} derived corresponding generalization bounds that depend on this margin rather than on input dimensionality. Because the decision of a nearest prototype classifier depends only on distances to a finite prototype set, the distance from a point to the decision boundary can be computed in closed form, yielding a robustness \emph{certificate} rather than an empirical estimate. Saralajew et al.~\cite{saralajew2020npc} made this explicit, showing that the hypothesis margin of nearest prototype classifiers is a tight lower bound on adversarial perturbation size and can be computed in constant time for seminorm-induced dissimilarities. GLVQ trained with triplet losses automatically maximizes this margin, certifying an entire test set in roughly the time required by randomized smoothing for a single CIFAR-10 sample. Voráček and Hein~\cite{Voracek2022} extended this analysis, demonstrating that $\ell_2$-PNPC (Provably Adversarially Robust Nearest Prototype Classifier) achieves the best certified $\ell_2$-robust accuracy on MNIST among deterministic certifiers, outperforming Saralajew et al.'s $\ell_2$-GLVQ by approximately 6\% robust accuracy at $\epsilon_2 = 1.58$. On CIFAR-10 under the LPIPS perceptual metric, their Perceptual-PNPC reaches 80.3\% clean accuracy with 40.5\% certified robust accuracy, compared to 71.6\% clean and only 9.8\% empirical robust accuracy for an adversarially trained ResNet-50 baseline. Empirically, Saralajew et al.~\cite{SaralajewGLVQ} found that GLVQ and its tangent-distance variant GTLVQ exhibit high \emph{base} robustness without any adversarial training, and that adversarial examples against these models resemble semantically meaningful interpolations towards the target class rather than arbitrary noise. 

This connection between margin and robustness is not unique to prototype-based classifiers. The linear Support Vector Machine (SVM) is a classic maximum-margin classifier, and its robustness properties have been extensively studied~\cite{elsayed2018margin, dakhale2023gradient}. Inspired by these results, Elsayed et al.~\cite{elsayed2018margin} proposed a margin-maximization objective for deep neural networks, demonstrating that it enhances robustness to adversarial perturbations by promoting a larger geometric margin. This approach effectively extends the classical SVM margin-maximization principle to the deep learning setting. Yu et al.~\cite{yu2019making} further showed that classical machine learning pipelines, including SVMs, can be integrated into neural network architectures and jointly optimized through backpropagation. Their findings highlight the flexibility of SVM-based approaches within modern deep learning frameworks. Motivated by this, we include a linear SVM as an additional baseline alongside linear and GLVQ classifiers, enabling a direct comparison between a conventional maximum-margin classifier and the prototype-based GLVQ approach. Related studies investigating margin-based robustness in non-parametric and nearest-neighbor classifiers have reported similar findings~\cite{wang2018analyzing, yang2020robustness}.

Unfortunately, shallow classifiers such as GLVQ or SVM tend to struggle in highly nonlinear settings such as image data, where the Euclidean distance in input space is uninformative for class discrimination~\cite{Khalid2025ContEx}, a phenomenon consistent with the accuracy--interpretability trade-off described in the XAI literature~\cite{gunning2850darpa, xu2019explainable}. This has motivated hybrid architectures that use a deep neural network to perform a nonlinear embedding while retaining prototype-based decisions in the embedding space. Examples are Prototypical Networks~\cite{snell2017prototypical}, autoencoder-plus-prototype-layer models whose prototypes decode back to the input space~\cite{li2018deep}, ProtoPNet and its part-prototype descendants~\cite{chen2019protopnet}, classification-by-components~\cite{saralajew2019cbc}, and prototype--rule hybrids~\cite{angelov2020dmr}. These hybrid models improve classification performance but lose robustness guarantees, because the certificate of a nearest prototype classifier is only valid in the space where the prototypes are defined: the \emph{embedding} space~\cite{Khalid2026, biehl2016wires}. An unconstrained embedding network can map an arbitrarily small input perturbation to an arbitrarily large embedding displacement, so a certified radius in $\Phi$ provides no information regarding robustness in $\mathcal{X}$ unless the embedding network's sensitivity is itself bounded. Bounding the embedding network's sensitivity is, therefore, the starting point for our method.
\section{Methodology}\label{sec:method}

Our goal is to leverage the interpretability and robustness advantages of prototype-based models when combined with deep embedding networks for pre-processing. We first establish the theoretical foundation of our method before describing our proposed algorithm for Lipschitz training.

\subsection{Adversarial Robustness}

Assume a deep neural network $\phi : \mathcal{X} \to \Phi$, which maps input data from some input space $\mathcal{X}$ to an $n$-dimensional embedding space $\Phi$. Further, assume that the input space $\mathcal{X}$ is equipped with some metric $d$ and the embedding space $\Phi$ with some metric $D$. Finally, assume that we have trained a classifier $f : \Phi \to \{1, \ldots, K\}$ which maps data in the embedding space to one of $K$ labels.

\begin{figure}
	\centering
	\includegraphics[width=8.5cm]{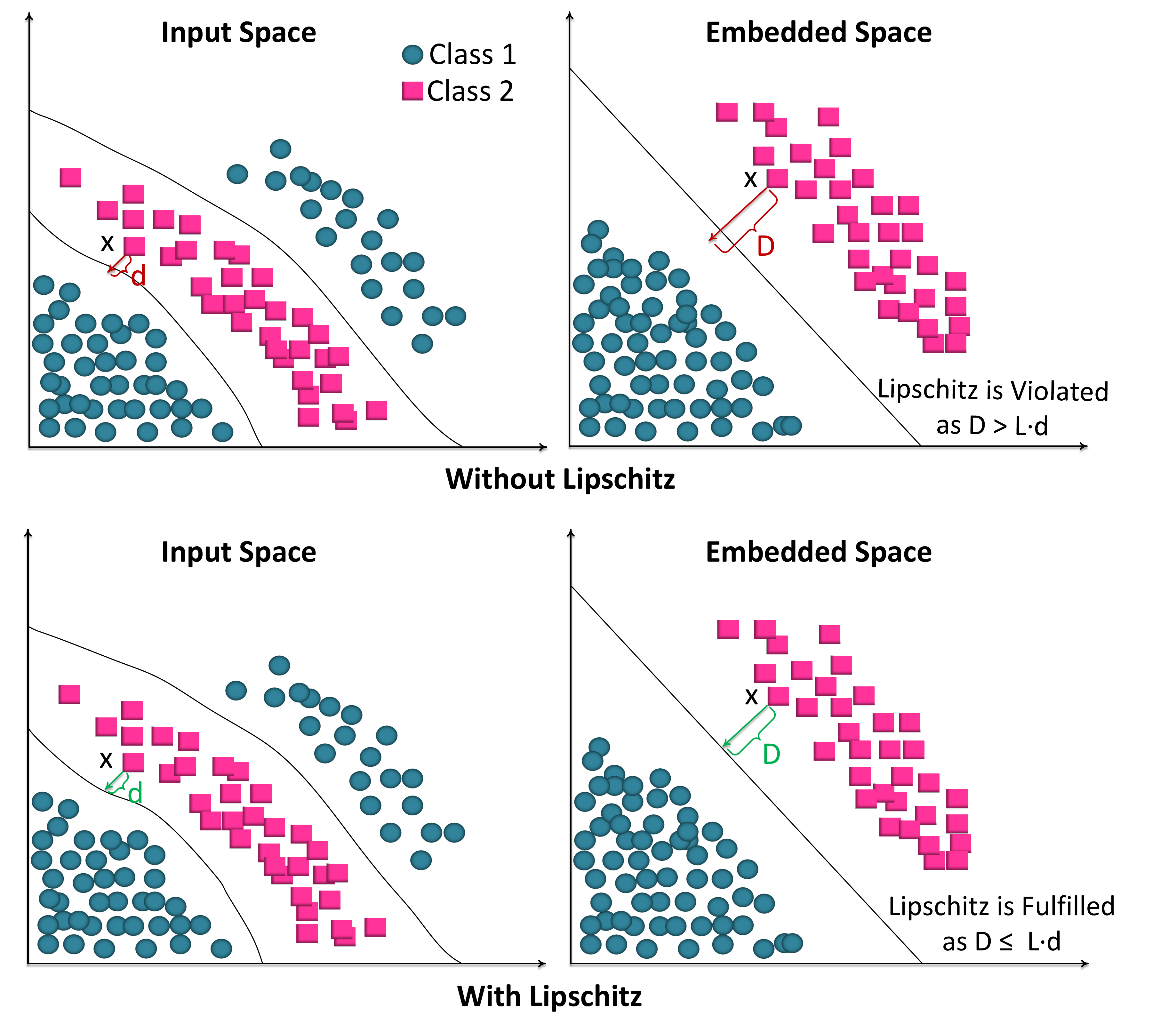}
	
	\caption{Top Left: An adversarial attack of size $d$ from point $x$. Top right: In embedding space, the attack is enlarged to size $D$, violating Lipschitz continuity. Bottom: With Lipschitz training, the classification boundary shifts such that the attack fails because $D$ is now upper-bounded by $L \cdot d$.}
	\label{fig:mainidea}
	
\end{figure}

The core challenge we try to address is that a deep neural network $\phi$ may yield an embedding space that improves classification accuracy but breaks adversarial robustness. Points that are close in the input space may be suddenly far apart in the embedding space. Hence, even if the classifier $f$ is adversarially robust in the embedding space, this does not guarantee adversarial robustness in the input space (see Figure~\ref{fig:mainidea}, top). Our core argument is straightforward: Lipschitz-continuity limits how much distances in the input space can be stretched in the embedding space (see Figure~\ref{fig:mainidea}, bottom). Hence, a robust classifier in the embedding space, in combination with a deep network that satisfies the Lipschitz property, yields a robust classifier in the input space.

To make our argument precise, we first introduce a formal definition of (local) adversarial robustness.

\begin{definition}
Let $\Phi$ be some set, let $f$ be some classifier $f : \Phi \to \{1, \ldots, K\}$ for some $K \in \N$, and let $D: \Phi \times \Phi \to \R$ be a metric on $\Phi$. Further, let $x \in \Phi$ and $\epsilon \in \R$ with $\epsilon > 0$. We say that $f$ is locally $\epsilon$-robust around $x$ (with respect to metric $D$) if and only if for all $y \in \Phi$ with $D(x, y) \leq \epsilon$ we have $f(x) = f(y)$.
\end{definition}

For the purpose of our argument, we assume that this property holds point-wise for all correctly classified points, in particular for the held-out points used for robustness evaluation. 
It is important to note that the robustness radius $\epsilon$ is not a fixed constant but depends on the input point $x$. For a nearest prototype classifier such as GLVQ, the certified radius around a correctly classified point $x$ is half the distance from $x$ to the decision boundary~\cite{crammer2002} induced by the closest correct and closest incorrect prototype. Let $d(x,w) = \lVert x - w \rVert$ denote the $\ell_2$ distance between $x$ and $w$. Saralajew et al.~\cite{saralajew2020npc} show that the \emph{hypothesis margin} at $x$ can be obtained in constant time during the inference step as:
	\begin{equation}
		\epsilon(x) \;=\; \tfrac{1}{2}\Big( \lVert \phi(x) - \phi(w^{-}) \rVert - \lVert \phi(x) - \phi(w^{+}) \rVert \Big),
		\label{eq:hypothesis_margin}
	\end{equation}
where $w^{+}$ and $w^{-}$ denote the closest prototype of the correct class and any incorrect class, respectively. This quantity lower-bounds the sample margin and hence the norm of any adversarial perturbation that changes the predicted label, with the bound being tight. The notion originates with Crammer et al.~\cite{crammer2002}, who introduced the hypothesis margin to derive generalization bounds for LVQ under the $\ell_2$ norm. The certificate applies only to correctly classified points; for misclassified points, the signed margin of~\cite{saralajew2020npc} is negative, and no radius is guaranteed.

Despite the robustness certificates of LVQ models, it is easy to construct a case where an LVQ model loses its robustness when combined with a deep embedding model. The certificate in Eq.~\eqref{eq:hypothesis_margin} is a statement about distances in the embedding space $\Phi$. If $\phi$ is unconstrained, an input perturbation of arbitrarily small norm in $\mathcal{X}$ can be mapped to a distance larger than $\epsilon(x)$ in $\Phi$, so the robustness guarantee in $\Phi$, by itself, carries no information about robustness in $\mathcal{X}$. Therefore, we also need Lipschitz continuity.

\subsection{Lipschitz continuity}

To ensure that robustness guarantees in the embedding space translate to robustness guarantees for a deep prototype classifier (including a deep embedding model), we require (local) Lipschitz continuity as a second ingredient.

\begin{definition}
Let $\mathcal{X}$ and $\Phi$ be sets, let $\phi : \mathcal{X} \to \Phi$, let $d : \mathcal{X} \times \mathcal{X} \to \R$ be a metric on $\mathcal{X}$, and let $D : \Phi \times \Phi \to \R$ be a metric on $\Phi$. Further, let $x \in \mathcal{X}$ and let $L \in \R$ with $L > 0$. We say that $\phi$ is locally $L$-Lipschitz-continuous around $x$ (with respect to metrics $d$ and $D$) if and only if for all $y \in \mathcal{X}$ we obtain
\begin{equation}
D(\phi(x), \phi(y)) \leq L \cdot d(x, y).
\end{equation}
\end{definition}

It follows that the local robustness of a classifier $f$, combined with the local Lipschitz continuity of an embedding model $\phi$, implies the local robustness of $f$ combined with $\phi$.

\begin{theorem}\label{thm:main}
Let $\mathcal{X}$ and $\Phi$ be sets, let $\phi : \mathcal{X} \to \Phi$, let $f : \Phi \to \{1, \ldots, K\}$ for some $K \in \N$, let $d : \mathcal{X} \times \mathcal{X} \to \R$ be a metric on $\mathcal{X}$, and let $D : \Phi \times \Phi \to \R$ be a metric on $\Phi$. Further, let $x \in \mathcal{X}$, and $\epsilon, L \in \R$ with $\epsilon > 0$ and $L > 0$.

Then it holds: if $\phi$ is locally $L$-Lipschitz around $x$ (w.r.t.\ metrics $d$ and $D$) and $f$ is locally $\epsilon$-robust around $\phi(x)$ (w.r.t.\ metric $D$), then $\phi \circ f$ (defined as $f(\phi(x))$) is $\frac{\epsilon}{L}$-robust around $x$ with respect to $d$.

\begin{proof}
Consider any $y \in \mathcal{X}$ with $d(x, y) \leq \frac{\epsilon}{L}$. Then, $D(\phi(x), \phi(y)) \leq L \cdot d(x, y) \leq \epsilon$ because $\phi$ is $L$-Lipschitz around $x$ with respect to $d$ and $D$. Hence, $f(\phi(x)) = f(\phi(y))$ because $f$ is $\epsilon$-robust around $\phi(x)$ w.r.t.\ $D$.
\end{proof}
\end{theorem}
Here $d$ and $D$ are the $\ell_2$ metrics on $\mathcal{X}$ and $\Phi$; the squaring in Equation~\eqref{eq:lipschitz_loss} applies inside the Lipschitz loss only, so that larger violations are punished more strongly.

While this theorem is, admittedly, somewhat trivial, it is highly useful for our purpose. The theorem implies that we can leverage the robustness and interpretability properties of GLVQ models in combination with deep networks if the deep networks are Lipschitz continuous with sufficiently small $L$. Hence, we devise a training method that encourages this Lipschitz property.

\subsection{Lipschitz Training}

Our goal is to construct a training scheme which ensures that a deep neural network $\phi$ is locally $L$-Lipschitz-continuous for some desired $L$ around all training data points $x$. To achieve this goal, our core strategy is quite simple: For each training data point $x_i$, we sample one neighbor $y_i$ by adding Gaussian random noise to $x_i$. Then, we punish violations of the Lipschitz property by imposing the loss following loss:
\begin{align}
\hspace{-1cm}
\mathcal{L}_\text{Lipschitz} &:= \sum_{i=1}^N \Big[ D(\phi(x_i), \phi(y_i))^2 - L^2 \cdot d(x_i, y_i)^2 \Big]_+,  \label{eq:lipschitz_loss} \\
&\text{where,}\quad y_i \sim \mathcal{N}(x_i, \sigma^2 \cdot \bm{I})
\end{align}
and where $[\cdot]_+$ denotes the hinge loss/the rectified linear unit $[\ell]_+ = \max\{0, \ell\}$ and $\mathcal{N}$ denotes the multivariate Gaussian density.

For training, the Lipschitz loss~\eqref{eq:lipschitz_loss} is added to the regular classification loss of the classifier. In our case, we train a deep embedding network $\phi$ with a final linear layer for classification using the cross-entropy loss
\begin{equation}
	\mathcal{L} = \mathcal{L}_{\text{CE}} + \lambda\,\mathcal{L}_{\text{Lipschitz}},
	\label{eq:loss_stage1}
\end{equation}
where $\lambda$ is the Lipschitz weight, controlling the strength of the Lipschitz loss relative to the cross-entropy loss.

The loss is then minimized via standard deep learning optimizers, such as Adam.

\paragraph{Hyperparameters:} Lipschitz training depends on three hyperparameters, each playing a distinct role.

The hyperparameter $L$ is the desired upper bound on the Lipschitz constant of the embedding network $\phi$. As established in Theorem \ref{thm:main}, if $\phi$ is $L$-Lipschitz and the classifier $f$ is $\epsilon$-robust in the embedding space, then the composed classifier is $\epsilon/L$-robust in the input space. Thus, $L$ enters the robustness radius directly through the factor $1/L$. It may, therefore, appear tempting to set $L$ as low as possible to achieve the best possible robustness bound. However, imposing tighter Lipschitz continuity on $\phi$ also means that distances within the embedding space shrink, which in turn reduces the distances from which the hypothesis margin in Equation~\ref{eq:hypothesis_margin} is computed. 
The radius $\epsilon(x)/L$ that Theorem~\ref{thm:main} transfers to the input space is therefore a ratio of two quantities that both shrink as $L$ shrinks, so it does not simply grow as $1/L$. Conversely, a larger $L$ relaxes the constraint, potentially improving accuracy but weakening the robustness guarantee. The choice of $L$ should therefore be guided by the desired perturbation budget in the input space.

The second hyperparameter is the Lipschitz weight $\lambda$, which controls how strongly violations are penalized relative to the classification loss. The higher $\lambda$ is, the more the Lipschitz loss is weighted and the fewer violations of the the Lipschitz loss are expected. However, higher $\lambda$ may also hurt classification accuracy (especially on non-attacked data) because the classification loss has less influence on training. $\lambda$ can thus be used to navigate the accuracy-robustness tradeoff.

The third hyperparameter is the standard deviation $\sigma$ of the isotropic, zero-mean Gaussian noise from which the neighbor $y_i$ for each training sample $x_i$ is generated. Lower values mean that neighbors are samples more closely to the training data, higher values mean that neighbors are farther away. We generally recommend to base $\sigma$ on the expected strength of adversarial attacks. 

All three hyperparameters are also investigated experimentally and discussed in Section~\ref{sec:hyperparams_selection}.

\paragraph{Computational Efficiency:} In terms of computational efficiency, Lipschitz training is releatively cheap. For every training point $x_i$ in a mini-batch, we sample a single neighbor $y_i$ via Gaussian noise. Thus, the asymptotic training efficiency remains unchanged. Lipschitz training compares particularly favorably to adversarial training~\cite{madry2017towards}, where $y_i$ needs to be generated via projected gradient descent, which requires an inner maximization involving several forward and backward passes for every single training data point during training. 

\paragraph{Lack of theoretical guarantees:} We acknowledge that Lipschitz training does \emph{not} guarantee a certain Lipschitz constant but only punishes violations of the desired constant $L$ for the distance between a training sample $x_i$ and the sampled neighbor $y_i$. If the Lipschitz loss~\eqref{eq:lipschitz_loss} becomes zero during training, $L$ does hold for the training data because then $\|\phi(x_i) - \phi(y_i)\|^2 \leq L^2 \|x_i - y_i\|^2$ for all $(x_i, y_i)$ pairs. Further, we recommend to re-sample the neighbors $y_i$ in each mini-batch, thus providing a more comprehensive stochastic sampling in each direction. If training would continue indefinitely under infinitely strong weight $\lambda$, local $L$-Lipschitz continuity would eventually result. In practice, though, the effective Lipschitz constant of $\phi$ will be higher than $L$. We investigate the empiric effectiveness of Lipschitz training in much more detail in the next section.

\subsection{Classification Heads}\label{sec:classification_heads}

After training the embedding network $\phi$, we use one of three classification heads for inference. By \enquote{classification head}, we mean the final layer $f$ that maps the embedding $\phi(x)$ to a class prediction.

\textbf{Linear:} The standard linear+softmax layer with cross-entropy loss serves as our baseline and is used during the first training stage.

\textbf{Linear SVM:} We extract the embeddings of the training fold with the embedding network frozen and fit a linear SVM on these fixed feature vectors using \texttt{scikit-learn}'s \texttt{LinearSVC} ($C = 1$, one-vs-rest). Since the decision function of a linear SVM, $f(z) = Wz + b$, has the same form as a linear layer, the learned $W$ and $b$ are copied into a linear layer on top of the frozen embedding network~\cite{yu2019making}, so that the same gradient-based attacks apply as for the other heads. This head provides a maximum-margin classifier in the embedding space but lacks interpretability.

\textbf{Generalized Learning Vector Quantization (GLVQ):}  A prototype-based model to achieve interpretable classification. The prototypes $w_1, \ldots, w_K$ are initialized in the input space and then mapped into $\Phi$ by the embedding network $\phi$, rather than being free vectors in $\Phi$. This keeps them interpretable as images throughout training and guarantees that they lie on the image of $\phi$, which makes the distances in embedding space and input space 
comparable between prototypes and data. The optimization of the prototypes occurs via Adam on the GLVQ loss \cite{sato1995generalized}. However, in contrast to regular GLVQ, because the prototypes live in $\mathcal{X}$, the GLVQ loss needs to be propagated through the embedding network $\phi$. Still, the parameters of $\phi$ remain frozen at this stage. 

Because both GLVQ and SVM are maximum margin classifiers, we assume that they are locally $\epsilon$-robust, such that the Lipschitz-continuous embedding network $\phi$ combined with GLVQ or SVM should be $\frac{\epsilon}{L}$-robust, following Theorem~\ref{thm:main}. We will test this hypothesis experimentally.

\subsection{Summary of Training Procedure}\label{sec:training_summary}
In summary, training proceeds in two stages. 
\textbf{Stage 1: Lipschitz Embedding Training.} We train the embedding network $\phi$ jointly with a linear (+softmax) classification head, minimizing Eq.~\eqref{eq:loss_stage1}. This step (hopefully) yields a locally $L$-Lipschitz continuous embedding network $\phi$. 

\textbf{Stage 2: Freeze and Fit Classifier Heads.} After Stage 1, we freeze the embedding network $\phi$, keeping it in evaluation mode and setting \texttt{requires\_grad = False}. The linear head is \emph{not} re-trained: we keep the one obtained in Stage 1. Only the SVM and the GLVQ prototypes are fitted, both on top of the frozen embedding, ensuring that all three heads share identical features and differ only in the decision rule. 
\section{Experimental Setup and Datasets}

Our primary objective is to test whether a Lipschitz-constrained embedding network combined with a maximum-margin classifier (GLVQ or SVM) achieves higher adversarial robustness than all baseline combinations. Specifically, we aim to:
\begin{enumerate}
\item isolate the effect of the Lipschitz loss by comparing standard and Lipschitz-constrained embedding networks with the same classification head;

\item  isolate the effect of the classification head by comparing Linear, SVM, and GLVQ heads on the same frozen embedding; and

\item demonstrate that the proposed Lipschitz-GLVQ model achieves enhanced robustness compared to all baselines, as measured by standard adversarial attacks.
\end{enumerate}

We evaluate the proposed scheme on three image classification benchmarks of increasing difficulty and resolution. Our aim is not to reach state-of-the-art clean accuracy but to investigate the effect of the Lipschitz loss: within each dataset, every model shares the same embedding network architecture, the same optimization budget, and the same data splits, so that differences in adversarial accuracy can be attributed to the Lipschitz loss and to the choice of classification head rather than to model capacity or tuning. 

\subsection{Datasets}\label{sec:datasets}
The detailed description of all three datasets is given below:
\subsubsection{MNIST}
The first benchmark is MNIST~\cite{lecun2010mnist}, consisting of $28\times 28$ gray-scale images (784 input features) with raw pixel values in $[0, 255]$. The images are loaded with \texttt{transforms.ToTensor()}, which maps them to $[0,1]$; no further channel normalization is applied. The \texttt{torchvision.datasets} release contains $70,000$ samples ($60,000$ train, $10,000$ test) evenly distributed over ten digit classes (0--9). The class distribution is approximately, but not exactly, uniform. 

\subsubsection{CIFAR-10}
The second benchmark is CIFAR-10~\cite{krizhevsky2009learning} (accessed via PyTorch), described by its authors as ``60000 32x32 color images in 10 classes, with 6000 images per class'', split into $50,000$ training and $10,000$ test images. The images are normalized with the channel mean and standard deviation computed on that split.

\subsubsection{Imagenette}
The third benchmark is Imagenette~\cite{howard2019imagenette}, described by Howard et al. as ``a subset of 10 easily classified classes from Imagenet''. We use the \texttt{imagenette-160} release with its predefined \texttt{train} / \texttt{val} directory structure (Imagenette-160\footnote{\url{https://github.com/fastai/imagenette}}), and resize all images to $64 \times 64$ pixels to keep the embedding network architecture consistent with the CIFAR-10 setting. We use the $9,469$ images in the \texttt{train} directory, which are normalized with the channel mean and standard deviation computed on that split.
\subsection{Model Architecture and Feature Extraction}
\textbf{Embedding network:} We consider image classification with a
convolutional neural network (CNN) for feature extraction. Detailed descriptions for each dataset and its respective settings for the CNN embedding network are provided in Table~\ref{tab:datasets}. The same embedding network architecture is used for both the standard (unregularized) models and the Lipschitz-constrained models, ensuring that any differences in performance can be attributed solely to the Lipschitz loss rather than architectural variations.
These architectures were chosen to be lightweight and representative of typical CNNs for each dataset. A systematic hyperparameter search over network architectures was outside the scope of this paper, and other architectures may yield better performance.

\begin{table*}[t]
	\centering
	\caption{Datasets, input pre-processing and network parameters. The pixel
		range refers to the values after \texttt{ToTensor} and before channel
		normalization.}
	\label{tab:datasets}
	\small
	\begin{tabular}{llll}
		\toprule
		& \textbf{MNIST} & \textbf{CIFAR-10} & \textbf{Imagenette-160} \\
		\midrule
		Classes                 & 10                & 10                & 10 \\
		Samples used (train)    & $60,000$        & $50,000$        & $9,469$ \\
		Native resolution       & $28\times28\times1$ & $32\times32\times3$ & variable (resized to $64\times64$) \\
		Input to the network    & $28\times28\times1$ & $32\times32\times3$ & $64\times64\times3$ \\
		Pixel range             & $[0, 1]$          & $[0, 1]$          & $[0, 1]$ \\
		Channel normalization   & none              & split mean/std    & split mean/std \\
		Convolutional layers    & 2                 & 3                 & 3 \\
		Batch normalization     & no                & yes               & yes \\
		Activation              & ReLU              & ReLU              & ReLU \\
		Pooling                 & $2\times2$ max    & $2\times2$ max    & $2\times2$ max \\
		Dropout                 & 0.0               & 0.5               & 0.5 \\
		Embedding $\dim(\Phi)$  & 16                & 64                & 64 \\
		\bottomrule
	\end{tabular}
\end{table*}

\begin{figure*}
	\centering
	\begin{tikzpicture}[
		pbox/.style = {
			draw, rounded corners=6pt,
			minimum height=1.4cm,
			minimum width=2.8cm,
			align=center,
			font=\small\bfseries,
			text=black
		},
		arr/.style = {-{Stealth[length=7pt]}, thick, black},
		]
		
		\node[pbox, fill=blue!15, draw=uniblue] (img)
		{Input Space $\mathcal{X}$};
		
		\node[pbox, fill=blue!25, draw=uniblue, right=20pt of img] (cnn)
		{CNN Embedding\\$\phi:\mathcal{X}\!\to\!\Phi$};
		
		\node[pbox, fill=blue!10, draw=uniblue, right=20pt of cnn] (emb)
		{Embedding Space $\Phi$\\metric $D$};
		
		\draw[arr] (img) -- (cnn);
		\draw[arr] (cnn) -- (emb);
		
		\draw[warnorange, very thick, dashed, rounded corners=10pt]
		($(img.north west)+(90pt, 14pt)$)
		rectangle
		($(emb.south east)+(10pt, -10pt)$);
		
		\node[warnorange, font=\small\bfseries, anchor=south]
		at ($(img.north west)!0.5!(emb.north east)+(1.6, 18pt)$)
		{Lipschitz constraint enforced here};
		
		\draw[arr, line width=1.04pt, black]
		($(img.south west)!0.5!(emb.south east)+(1.6, -10pt)$)
		-- ++(0, -20pt);
		\node[pbox, fill=green!15, draw=okgreen,
		below=35pt of cnn]
		(glvq) {Classifier\\$f:\Phi\!\to\!\{1,\ldots,K\}$};
		
		\node[pbox, fill=green!35, draw=okgreen, right=29pt of glvq]
		(pred) {Predicted\\Class $\hat{y}$};
		
		\draw[arr] (glvq) -- (pred);
		
		\draw[okgreen, very thick, dashed, rounded corners=10pt]
		($(glvq.north west)+(-0.2, 4pt)$)
		rectangle
		($(pred.south east)+(4.2pt, -7pt)$);
		
		\node[okgreen, font=\small\bfseries, anchor=north]
		at ($(glvq.south west)!0.5!(pred.south east)+(0.4, -12pt)$)
		{Robustness certified here};
		
	\end{tikzpicture}
	\captionof{figure}{The proposed architecture where the Lipschitz loss is applied to the embedding network (CNN) $\phi$, which is then frozen; after that, the classification head classifies in the embedding space $\Phi$.}\label{fig:architecture}
\end{figure*}
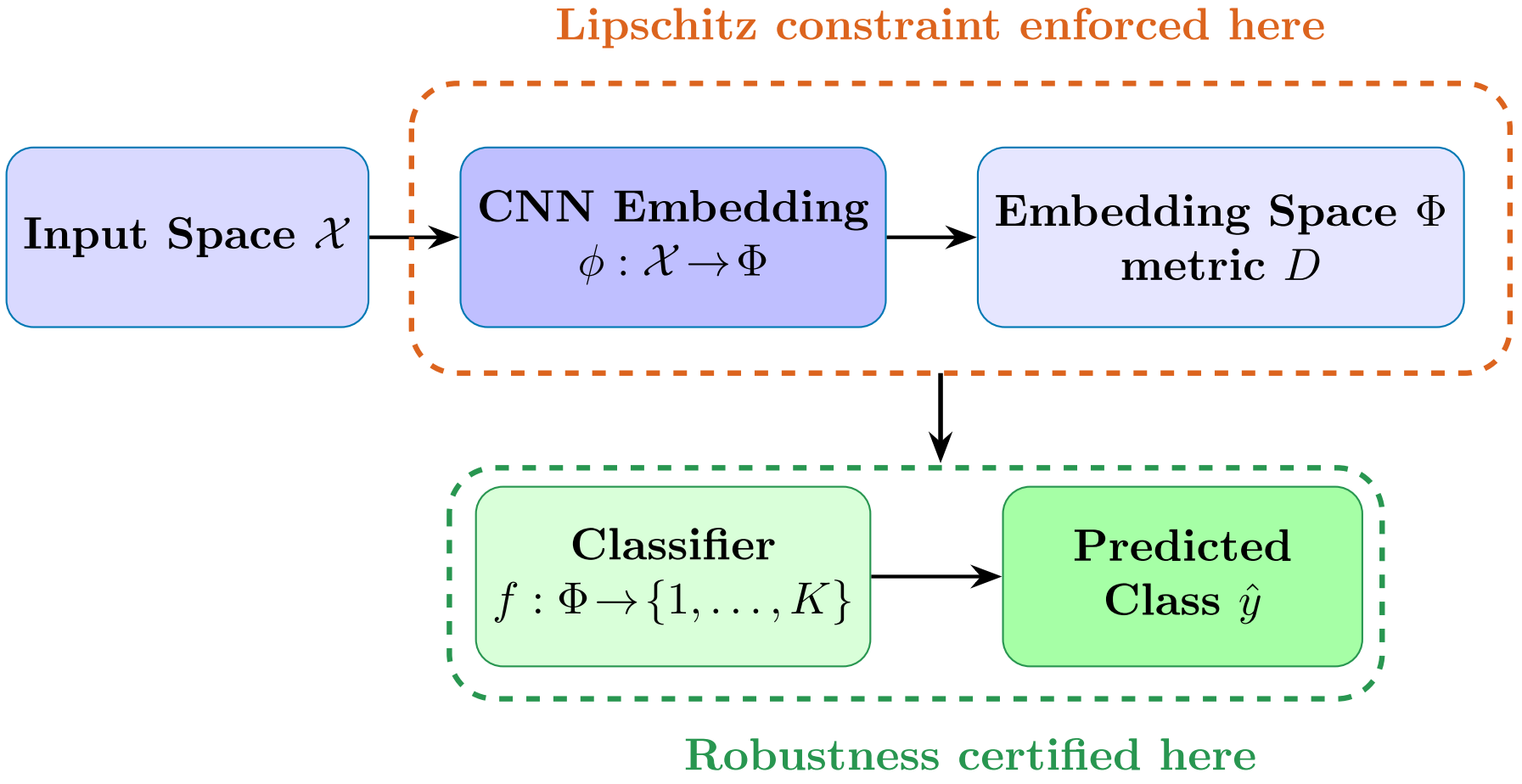

\textbf{Models' Architecture:} We compare six models, obtained by crossing three classification heads (linear, SVM, and GLVQ) with two training schemes for the embedding network (with and without the Lipschitz loss), as listed in Table~\ref{tab:models}. The training procedure consists of two stages. The embedding network is first trained end-to-end together with a linear (+softmax) head, with the Lipschitz loss of Eq.~\eqref{eq:loss_stage1} in the Lipschitz-constrained setting and without it in the baseline setting, and is then frozen, as shown in Figure~\ref{fig:architecture}. The SVM and GLVQ heads are subsequently fitted on top of this frozen embedding.
Under this design, any difference between Linear and Lipschitz-Linear can be attributed to the Lipschitz loss alone, and any difference between Lipschitz-Linear and Lipschitz-GLVQ is caused by the classification head alone. Consequently, this setup allows us to isolate the effect of the Lipschitz loss from that of the classification head. We expect that the two Lipschitz variants with a margin-based heads ({Lipschitz-SVM} and {Lipschitz-GLVQ}) will gain substantial adversarial robustness, as they combine adversarial robustness in the embedding space (due to maximum margin properites) with the Lipschitz continuity of the embedding network. The added value of Lipschitz-GLVQ over Lipschitz-SVM should only lie in additional interpretability due to the prototype-based classification scheme.
\begin{table*}[t]
	\centering
	\caption{The six models compared on every dataset. All heads within each configuration share the same frozen embedding network.}
	\label{tab:models}
	\small
	\begin{tabular}{llp{6.0cm}c}
		\toprule
		Model & Embedding Net & Classification head & Interpretable? \\
		\midrule
		\textsc{Linear}            & standard  & linear+softmax layer, cross-entropy loss & no \\
		\textsc{SVM}               & standard  & linear SVM ($C=1$, one-vs-rest) fitted on the frozen embedding and converted into an equivalent linear layer so that gradient-based attacks apply & no \\
		\textsc{GLVQ}              & standard  & one prototype per class, GLVQ cost with the sigmoid link; only the prototypes are updated & yes \\
		\midrule
		Lipschitz\_Linear  & Lipschitz & as \textsc{Linear} & no \\
		Lipschitz\_SVM     & Lipschitz & as \textsc{SVM} & no \\
		\textbf{Lipschitz\_GLVQ} & \textbf{Lipschitz} & \textbf{as \textsc{GLVQ}} & \textbf{yes} \\
		\bottomrule
	\end{tabular}
\end{table*}

\subsection{Training Protocol}
The following training details apply to all experiments.
Models were trained for 50 epochs on all three datasets to ensure that no model receives a larger optimization budget than another. The batch sizes are $32$ for MNIST, $128$ for CIFAR-10, and $64$ for Imagenette-160. For GLVQ, we use one prototype per class, as described in Section~\ref{sec:classification_heads}. The prototypes are the only parameters updated during GLVQ training: the embedding network is frozen, and the optimizer is constructed over the prototypes only. Prototypes are initialized as one randomly drawn training image per class and are updated directly in input space, so that they remain interpretable as images.
The embedding networks and linear heads are trained jointly in Stage 1 with an Adam optimizer ($\text{lr}=10^{-3}$, weight decay $10^{-4}$) for $50$ epochs; the GLVQ prototypes are trained with a separate Adam instance at the same learning rate. The SVM is fitted with \texttt{scikit-learn} rather than by gradient descent, as described in Section~\ref{sec:classification_heads}. The remaining hyper-parameter settings are listed in Table~\ref{tab:training}.

\begin{table*}[t]
	\centering
	\caption{Training hyperparameters per dataset.}
	\label{tab:training}
	\small
	\begin{tabular}{lccc}
		\toprule
		& MNIST & CIFAR-10 & Imagenette \\
		\midrule
		Epochs                          & $50$   & $50$   & $50$ \\
		Batch size                      & $32$   & $128$  & $64$ \\
		Optimiser                       & \multicolumn{3}{c}{Adam, $\text{lr}=10^{-3}$, weight decay $10^{-4}$} \\
		Cross-validation folds          & $5$    & $5$    & $5$ \\
		Prototypes per class $P$        & $1$    & $1$    & $1$ \\
		Embedding dimension $n$         & $16$   & $64$   & $64$ \\
		\midrule
		\multicolumn{4}{l}{\textit{Lipschitz loss}} \\
		Neighborhood mean $\mu$        & $0$    & $0$    & $0$ \\
            Neighborhood std.\ dev.\ $\sigma$ & $0.1$ & $0.03$ & $0.03$ \\
		Lipschitz constant $L$             & $0.032$ & $0.032$ & $0.032$ \\
		Lipschitz weight $\lambda$        & $10$ & $10$ & $10$ \\
		\midrule
		\multicolumn{4}{l}{\textit{Attack parameters}} \\
		FGSM/PGD budget $\epsilon$      & $[0,\,0.4]$, 41 pts & $[0,\,0.0667]$, 35 pts & $[0,\,0.0667]$, 35 pts \\
		Evaluation batch size           & $200$ & $200$ & $200$ \\
		PGD steps $T$                   & $40$ & $20$ & $20$ \\
		PGD random restarts             & $1$ & $1$ & $1$ \\
		C\&W binary-search steps $N$    & $9$ & $9$ & $9$ \\
		C\&W optimisation steps $M$     & $1000$ & $1000$ & $1000$ \\
		C\&W initial constant           & $1$ & $1$ & $1$ \\
		C\&W confidence $\kappa$        & $0$ & $0$ & $0$ \\
		\bottomrule
	\end{tabular}
\end{table*}
\subsection{Lipschitz Hyperparameter Selection} \label{sec:hyperparams_selection}
As discussed previously in Section~\ref{sec:method}, Lipschitz training has three hyperparameters: the desired Lipschitz constant $L$, the Lipschitz weight $\lambda$, and the neighborhood standard deviation $\sigma$. We selected these values on a held-out portion of the training split (using only the official training split), keeping them fixed across all cross-validation folds.

\textbf{Lipschitz constant $L$.} As discussed in Section~\ref{sec:method}, $L$ has two opposing effects. On the one hand, Theorem~\ref{thm:main} certifies a radius of $\epsilon/L$ in input space, so a smaller $L$ (if actually achieved) means a larger guaranteed radius. On the other hand, a smaller $L$ constrains the embedding network $\phi$ more strongly, which may cost clean accuracy. The choice of $L$ therefore has to be made empirically.
We tested $L \in \{0,\, 0.032,\, 0.071,\, 0.1,\, 0.224,\, 0.316,\, 0.707,\, 0.775,\, 1\}$ (see Figure~\ref{fig:comparison_l}). Values of $L \gtrsim 0.707$ left the constraint inactive, while $L \leq 0.032$ produced no additional robustness gain. We therefore chose $L = 0.032$ as a conservative operating point, balancing robustness with (unattacked) classification accuracy.

\textbf{Lipschitz weight $\lambda$.} $\lambda$ is purely an optimization-side quantity; it weighs the Lipschitz loss versus the classification loss. We tested $\lambda \in \{1, 5, 10, 20, 50, 100\}$ (see Figure~\ref{fig:comparison_w}). Higher values reduce the rate of violations against the Lipschitz constraint but cost clean accuracy (i.e. accuracy without adversarial attacks) on CIFAR-10 and Imagenette-160 (Table~\ref{tab:violation}). 
We selected $\lambda = 10$ as the default Lipschitz weight.

\textbf{Neighborhood standard deviation $\sigma$.} The parameter $\sigma$ fixes the radius at which neighbors are sampled during training for testing violations of the Lipschitz condition. To effectively check the Lipschitz constraints for small distances where adversarial attacks are expected, $\sigma$ should be of the same order as the largest attack budget $\epsilon_{\max}$ evaluated on that dataset. Guided by this, we swept $\sigma \in \{0.03, 0.07, 0.1, 0.2, 0.3, 0.5\}$ and selected $\sigma = 0.1$ for MNIST and $\sigma = 0.03$ for CIFAR-10 and Imagenette-160, which are the smallest values at which the Lipschitz loss remains clearly active throughout training; Section~\ref{sec:comparison} discusses this sweep in detail.

\subsection{Evaluation Protocol}\label{sec:protocol}
All experiments use $5$-fold cross-validation over the training split, with folds generated by a shuffled \texttt{KFold} split under a fixed seed. In each fold, all six models are trained on four folds, and the held-out fold is used for both clean and adversarial evaluation. All results are averaged over the five folds and reported as the mean $\pm$ one standard deviation. All hyperparameters were fixed before running any experiments (see Section~\ref{sec:hyperparams_selection}).

\subsubsection{Implementation details}
The Lipschitz loss is computed on the same mini-batch as the classification loss, using the pre-logit embedding. As discussed above, the SVM and GLVQ classification heads are fitted on top of the frozen embedding network; therefore, the Lipschitz loss plays no role in their training. Mini-batches are drawn by reshuffling the training fold at the start of every epoch. Perturbed samples used for the Lipschitz loss are not clipped to the valid pixel range, since the local Lipschitz condition is meant to hold in the ambient space and clipping would bias the sampled neighborhood at the boundary of $[0,1]$.
In addition to optimization and hyper-parameter tuning, a crucial implementation detail concerns the handling of input normalization. Because the networks operate on channel-normalized inputs while perturbation budgets are only meaningful in raw pixel units, every attack follows the same three-step convention: the image is un-normalized into raw pixel space, the adversarial attack is applied in pixel space, enforcing the attack budget $\epsilon$ as well as the bounds of the pixel range $[0, 1]$. Finally, the result is re-normalized before being fed to the model. 
As loss functions for adversarial attacks, cross-entropy is used for Linear and SVM classifiers and the GLVQ loss for GLVQ models. For the Lipschitz-GLVQ model, the attack uses the classification part of the cost only, without the Lipschitz loss term, since the Lipschitz loss is not part of the decision rule and including it would provide the attacker with a gradient that does not correspond to the model's predictions.

In preliminary experiments with $L = 1$, we mistakenly included the Lipschitz loss in the loss that was attacked. Because the Lipschitz loss directly works against adversarial attacks, this made the adversarial robustness appear much better than it actually was. Therefore, we caution other researchers against this mistake.  
\subsubsection{Attack schemes and robustness evaluation}
We evaluate the adversarial robustness by measuring the accuracy under FGSM~\cite{goodfellow2014explaining}, PGD~\cite{madry2017towards}, and C\&W $\ell_2$ attacks~\cite{carlini2017towards}. All three types of attack are computed on the held-out fold of the current cross-validation split. FGSM and PGD are run across a wide range of attack strengths (in line with the robustness curve proposal of~\cite{Goepfert2020b}), namely $\epsilon \in [0, 0.4]$ in steps of $0.01$ for MNIST, and $\epsilon = k/510$ for $k = 0,\ldots,34$, i.e.\ $\epsilon \in [0, 0.067]$, for CIFAR-10 and Imagenette-160. Regarding the attack configurations, FGSM and PGD are both $\ell_\infty$ attacks. PGD uses $T$ steps with step size $\alpha = 2.5\,\epsilon/T$ and a random start inside the $\epsilon$-ball, while the C\&W attack uses a confidence of $\kappa = 0$, $N$ binary-search steps over the trade-off constant, and $M$ optimization steps per search step. The values of $T$, $N$, $M$, and the number of random restarts are listed in Table~\ref{tab:training}. The C\&W attack is unbounded and is reported as a mean $\ell_2$ distortion rather than as an accuracy curve.

\section{Experimental Results}\label{sec:results}
We evaluate our approach on MNIST, CIFAR-10, and Imagenette-160 discussed in section \ref{sec:datasets}. For each dataset, we compare six models: three heads (Linear, SVM, GLVQ) crossed with two embedding networks (standard and Lipschitz-constrained) (see Section~\ref{sec:method}). Our goal is to separate the effect of the Lipschitz loss from the effect of the classification head. We report clean accuracy and adversarial accuracy under FGSM, PGD, and C\&W attacks. Unless stated otherwise, all results use $L = 0.032$ and Lipschitz weight $\lambda = 10$ (Section~\ref{sec:hyperparams_selection}).
All curves reported in Figures~\ref{fig:mnist_fgsm} -- \ref{fig:imnet_cw} are the means over 5-fold cross-validation, with error bars representing the standard deviation across folds.

\begin{table}[ht]
	\centering
	\caption{Clean accuracy (\%). Values are the mean $\pm$ standard deviation over 5 folds.}
	\label{tab:clean}
    \resizebox{\linewidth}{!}{%
    	\begin{tabular}{lccc}
    		\toprule
    		Model & MNIST & CIFAR-10 & Imagenette-160 \\
    		\midrule
                Linear (standard) & $98.24 \pm 0.10$ & $78.97 \pm 0.49$ & $73.20 \pm 2.34$ \\
                SVM (standard)    & $98.34 \pm 0.13$ & $78.79 \pm 0.60$ & $78.14 \pm 1.06$ \\
                GLVQ (standard)   & $97.82 \pm 0.16$ & $79.18 \pm 0.53$ & $76.40 \pm 1.15$ \\
                \midrule
                Lipschitz-Linear  & $98.36 \pm 0.10$ & $76.31 \pm 0.99$ & $65.19 \pm 2.24$ \\
                Lipschitz-SVM     & $98.46 \pm 0.10$ & $76.82 \pm 0.93$ & $68.35 \pm 1.12$ \\
                Lipschitz-GLVQ    & $98.09 \pm 0.17$ & $76.49 \pm 1.20$ & $67.19 \pm 1.16$ \\
    	
    		\bottomrule
    	\end{tabular}
    }
\end{table}
The effect of the Lipschitz loss on clean accuracy depends strongly on the dataset (Table~\ref{tab:clean}). On MNIST, the two Lipschitz models slightly exceed their unconstrained counterparts, while on CIFAR-10 the standard models lead. The Lipschitz loss gains $0.12$ percentage points on MNIST and costs $2.66$ on CIFAR-10, consistent with the known robustness--accuracy trade-off~\cite{tsipras2019robustness, zhang2019trades}. On Imagenette-160, the cost is largest, $8.01$ points for the Linear models ($65.19\%$ vs.\ $73.20\%$), reflecting the slower convergence of the constrained embedding network under the matched 50-epoch budget, as shown by the loss curves in Section~\ref{sec:comparison}.

The clean accuracy (without any adversarial attack at $\epsilon = 0$) of the GLVQ models are close to that of Linear and SVM on every dataset. GLVQ is $0.42$ percentage points below Linear on MNIST and $1.74$ points below SVM on Imagenette-160, while on CIFAR-10 the three heads are within $0.4$ points of each other.
Vor\'a\v{c}ek and Hein~\cite{Voracek2022} similarly report $97.3\%$ clean accuracy for $\ell_2$-PNPC on MNIST. Zhang et al.~\cite{zhang2019trades} observed a comparable trade-off on CIFAR-10, where TRADES ($1/\lambda = 6$) achieved $84.92\%$ clean accuracy against $56.61\%$ robust accuracy under PGD$_{20}$. Our clean accuracy is well below this, as expected given our relatively small embedding network with only $64$ embedding dimensions and 3 layers; we cite the comparison to situate the size of the trade-off, not to claim equality.

\subsection{Adversarial Robustness}
\begin{figure}
	\centering
	\begin{subfigure}[b]{0.49\textwidth}
		\includegraphics[width=\textwidth]{figs/FGSM_MNIST_fold.pdf}
		\caption{FGSM robustness on MNIST}\label{fig:mnist_fgsm}
	\end{subfigure}
	\hfill
	\begin{subfigure}[b]{0.49\textwidth}
		\includegraphics[width=\textwidth]{figs/PGD_MNIST_fold.pdf}
		\caption{PGD robustness on MNIST}\label{fig:mnist_pgd}
	\end{subfigure}
	\\[1ex]
	\begin{subfigure}[b]{0.49\textwidth}
		\includegraphics[width=\textwidth]{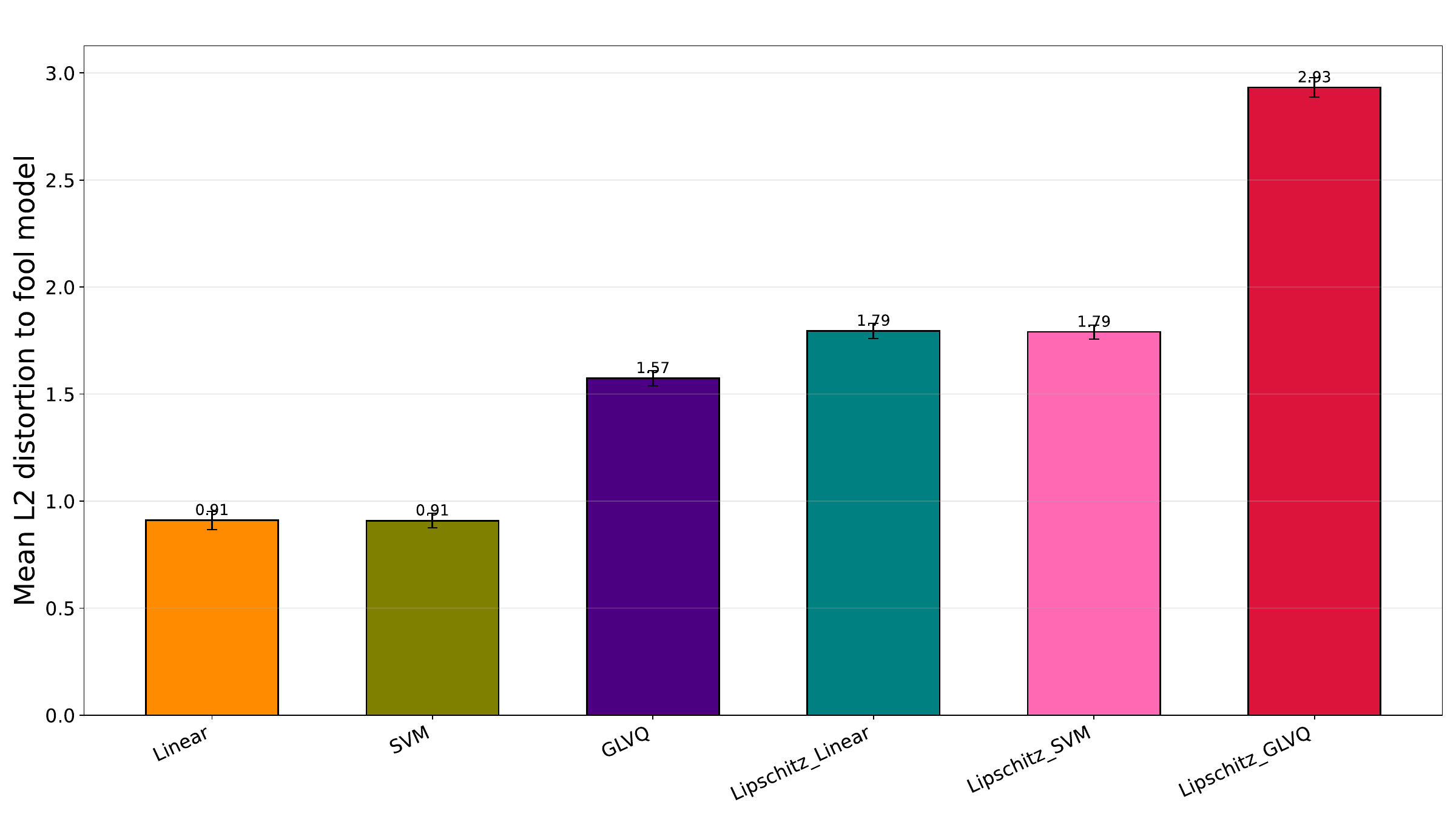}
		\caption{C\&W $\ell_2$ robustness on MNIST}\label{fig:mnist_cw}
	\end{subfigure}
	\caption{MNIST robustness. (\ref{fig:mnist_fgsm},\ref{fig:mnist_pgd}) Accuracy versus the $\ell_\infty$ attack budget $\epsilon$. (\ref{fig:mnist_cw}) Mean $\ell_2$ distortion required to flip the label; higher is more robust.}
	\label{fig:mnist}
\end{figure}

\begin{figure}
	\centering
	\begin{subfigure}[b]{0.49\textwidth}
		\includegraphics[width=\textwidth]{figs/FGSM_CIFAR_fold.pdf}
		\caption{FGSM robustness on CIFAR-10}\label{fig:cifar_fgsm}
	\end{subfigure}
	\hfill
	\begin{subfigure}[b]{0.49\textwidth}
		\includegraphics[width=\textwidth]{figs/PGD_CIFAR_fold.pdf}
		\caption{PGD robustness on CIFAR-10}\label{fig:cifar_pgd}
	\end{subfigure}
	\\[1ex]
	\begin{subfigure}[b]{0.49\textwidth}
		\includegraphics[width=\textwidth]{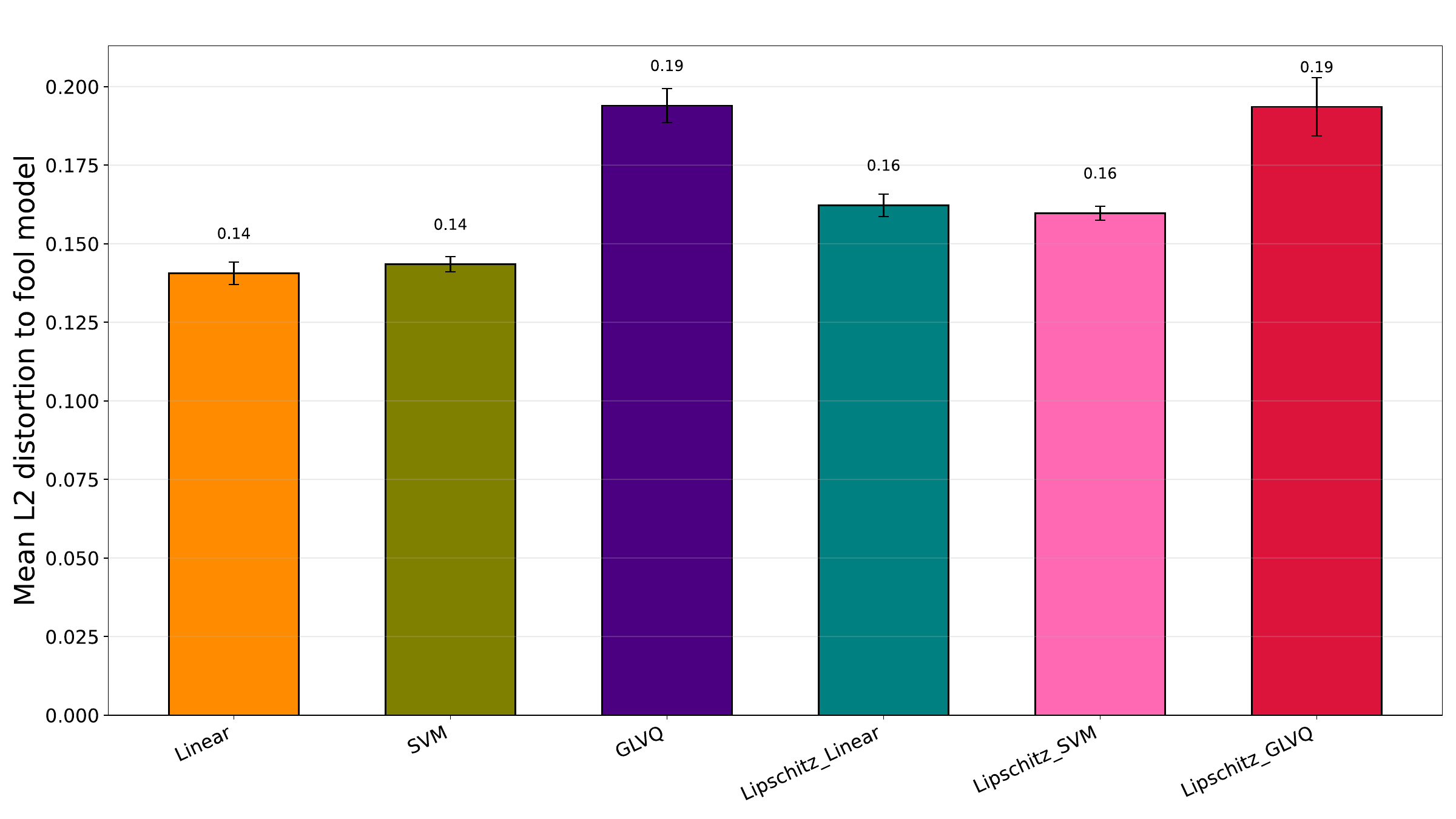}
		\caption{C\&W $\ell_2$ robustness on CIFAR-10}\label{fig:cifar_cw}
	\end{subfigure}
	\caption{CIFAR-10 robustness. (\ref{fig:cifar_fgsm},\ref{fig:cifar_pgd}) Accuracy versus the $\ell_\infty$ attack budget $\epsilon$. (\ref{fig:cifar_cw}) Mean $\ell_2$ distortion required to flip the label.}
	\label{fig:cifar10}
\end{figure}

At $\epsilon = 0$ (clean accuracy) all models achieve high accuracy. As $\epsilon$ increases, performance degrades rapidly for the standard models. Lipschitz-GLVQ achieves the highest robustness on MNIST and CIFAR-10 (Figures~\ref{fig:mnist_fgsm}--\ref{fig:cifar_cw}). On Imagenette-160 the picture differs: under PGD, Lipschitz-SVM is the strongest model over the range where accuracies remain substantial, with Lipschitz-GLVQ second, while under FGSM attacks, Lipschitz-GLVQ leads from $\epsilon \approx 0.016$ onwards by a wide margin ($10.8\%$ against $3.7\%$ at the largest budget). Among the unconstrained models, GLVQ is the most robust under FGSM attacks on all three datasets. This is likely because our embedding network already exhibits some Lipschitz continuity even without explicit training, while GLVQ's margin-based decision rule provides a natural robustness certificate in the embedding space~\cite{hammer2005, saralajew2020npc}.

\begin{figure}
	\centering
	\begin{subfigure}[b]{0.49\textwidth}
		\includegraphics[width=\textwidth]{figs/FGSM_Imagenette.pdf}
		\caption{FGSM robustness on Imagenette-160}\label{fig:imnet_fgsm}
	\end{subfigure}
	\hfill
	\begin{subfigure}[b]{0.49\textwidth}
		\includegraphics[width=\textwidth]{figs/PGD_Imagenette.pdf}
		\caption{PGD robustness on Imagenette-160}\label{fig:imnet_pgd}
	\end{subfigure}
	\\[1ex]
	\begin{subfigure}[b]{0.49\textwidth}
		\includegraphics[width=\textwidth]{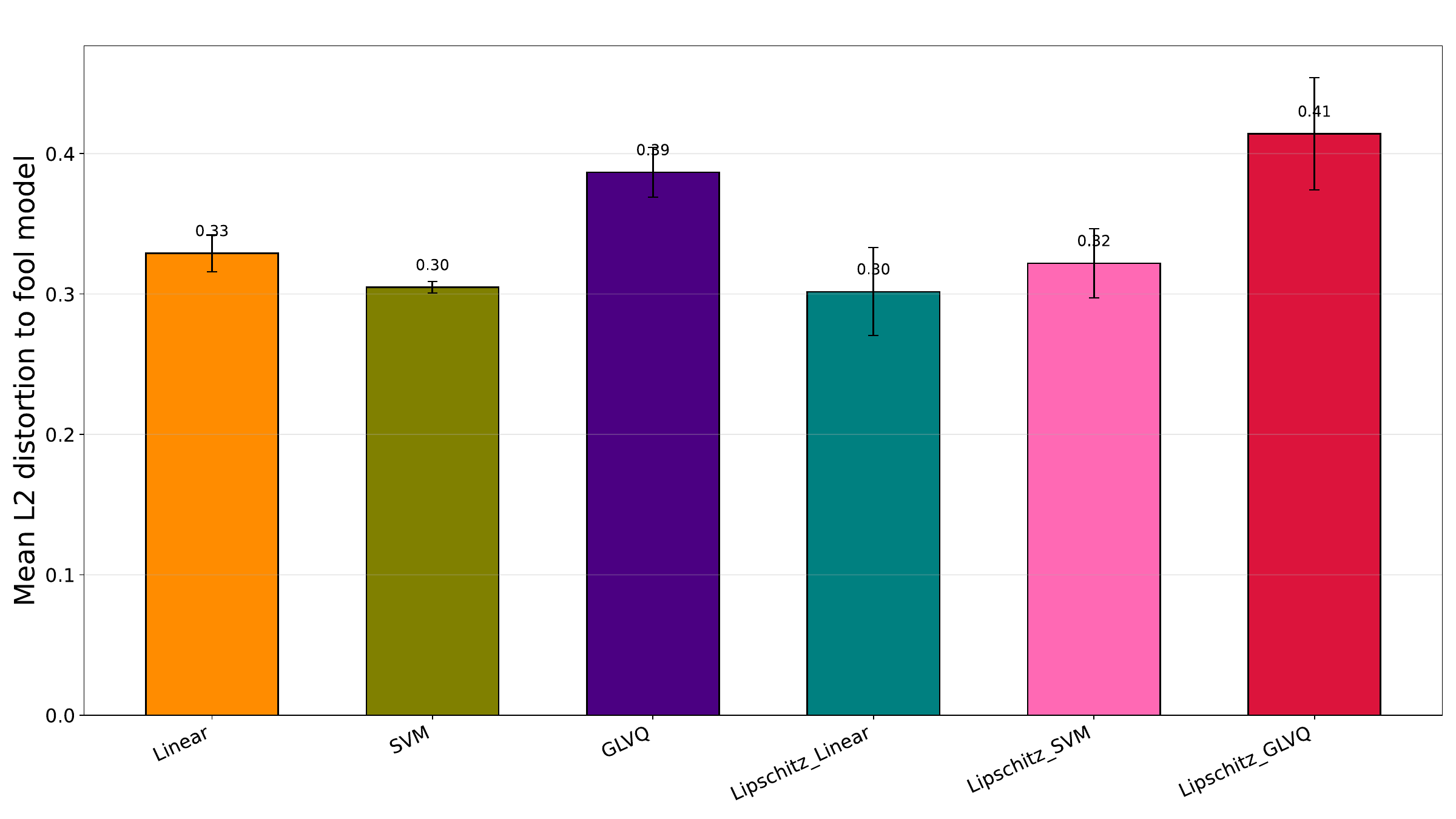}
		\caption{C\&W $\ell_2$ robustness on Imagenette-160}\label{fig:imnet_cw}
	\end{subfigure}
	\caption{Imagenette-160 robustness. (\ref{fig:imnet_fgsm},\ref{fig:imnet_pgd}) Accuracy versus the $\ell_\infty$ attack budget $\epsilon$. (\ref{fig:imnet_cw}) Mean $\ell_2$ distortion required to flip the label; higher values indicate greater robustness.}
	\label{fig:imagenet160}
\end{figure}

A key observation is that the improvement of Lipschitz-Linear and Lipschitz-SVM over their standard counterparts is strongest on MNIST. On CIFAR-10 (Figure~\ref{fig:cifar_fgsm}) and Imagenette-160 (Figure~\ref{fig:imnet_fgsm}), both also clearly outperform their unconstrained counterparts at small attack budgets. This advantage diminishes as $\epsilon$ increases, and on Imagenette-160 and under FGSM attacks on CIFAR-10, \emph{unconstrained} GLVQ overtakes them at higher attack strengths.

The results on MNIST are the clearest: under FGSM attacks (Figure~\ref{fig:mnist_fgsm}), Lipschitz-GLVQ retains $\approx 81\%$ accuracy at $\epsilon = 0.20$ and $\approx 43\%$ at $\epsilon = 0.40$. Under the stronger PGD adversary (Figure~\ref{fig:mnist_pgd}), it retains $\approx 53\%$ at $\epsilon = 0.20$, while every unconstrained model has collapsed to near-zero by $\epsilon \approx 0.19-0.20$. Degradation occurs only when the images are altered so strongly that a human observer might not recognize them, which is consistent with the goal of interpretability: a model should change its prediction only when the input has changed enough to change a human's judgment~\cite{vadillo2025adversarial}. On CIFAR-10 and Imagenette-160, however, a human observer would likely not change their predicted class even at the largest $\epsilon$ tested, so the current level of robustness is still insufficient for true interpretability, despite the significant improvement over the baseline.

The C\&W results (Figs.~\ref{fig:mnist_cw}, \ref{fig:cifar_cw}, \ref{fig:imnet_cw}) address the same question from a minimum-distortion perspective. On MNIST, the mean $\ell_2$ distortion needed to flip a label rises from $1.57$ to $2.93$ for the GLVQ pair ($1.87\times$) and from $0.91$ to $1.79$ for both Linear and SVM. On CIFAR-10 the two GLVQ variants are indistinguishable ($0.194$ against $0.194$), so Lipschitz training brings no measurable C\&W advantage there. On Imagenette-160, the ordering is favorable but within one standard deviation ($0.387 \rightarrow 0.414$). On CIFAR-10 and Imagenette-160, both prototype models require clearly more distortion than the four linear heads; on MNIST this holds for Lipschitz-GLVQ ($2.93$), while standard GLVQ ($1.57$) sits below the two Lipschitz linear heads.

In comparison to prior work, Vor\'a\v{c}ek and Hein~\cite{Voracek2022} report that $\ell_2$-PNPC attains $96.2\%$ certified accuracy on MNIST at $\epsilon_1 = 1$, outperforming $\ell_2$-GLVQ ($94.2\%$) and 1-NN ($95.0\%$). 
Our empirical results align with these findings, though our model is not certified in the same sense, because we enforce local Lipschitz continuity of the embedding only stochastically via training rather than proving a strict, global bound on the whole model. Saralajew et al.~\cite{SaralajewGLVQ} found that GLVQ and GTLVQ exhibit high base robustness without adversarial training, and that adversarial examples against them resemble semantically meaningful interpolations toward the target class rather than arbitrary noise; our observation that standard GLVQ beats Linear and SVM in terms of robustness supports this.

These results give partial support to our hypothesis. Lipschitz training on its own is clearly not sufficient: with a softmax head it buys almost nothing beyond MNIST, which is consistent with the observation that the most Lipschitz-stable model is the useless constant function~\cite{zuhlke2025adversarial}. The broader picture on the margin-based heads is more mixed than the theory would suggest. When comparing Lipschitz training to the choice of classifier, GLVQ appears to yield a clearer effect: unconstrained GLVQ already beats Linear and SVM everywhere, the SVM benefits much less from Lipschitz training compared to GLVQ, and on Imagenette-160 Lipschitz-GLVQ does not beat regular GLVQ. We return to these two discrepancies in Section~\ref{sec:discussion}.

\subsection{Hyperparameter Sensitivity Analysis}\label{sec:comparison}
\begin{figure}
	\centering
	\begin{subfigure}[b]{0.48\textwidth}
		\includegraphics[width=\textwidth]{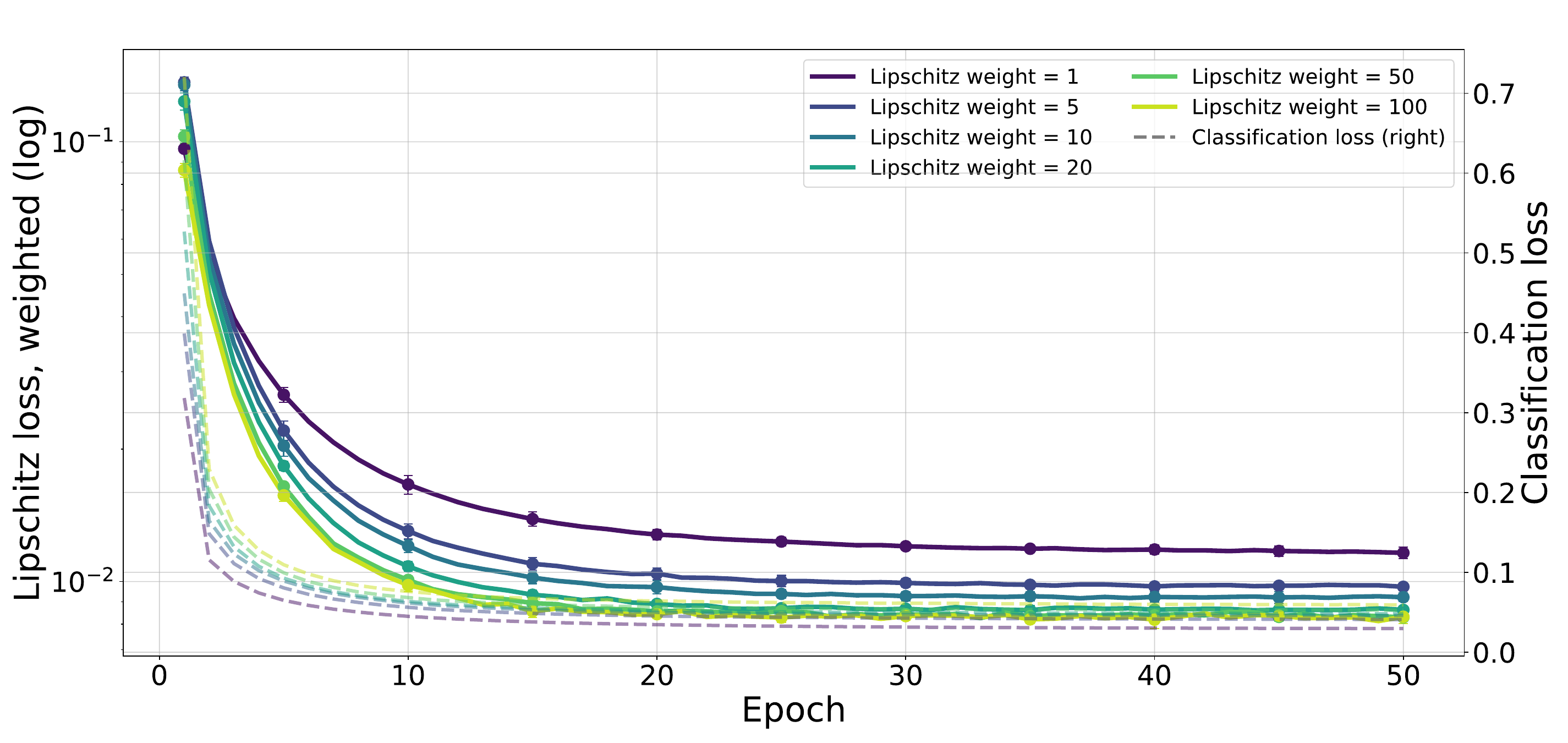}
		\caption{MNIST}\label{fig:comparison_w_mnist}
	\end{subfigure}
	\hfill
	\begin{subfigure}[b]{0.48\textwidth}
		\includegraphics[width=\textwidth]{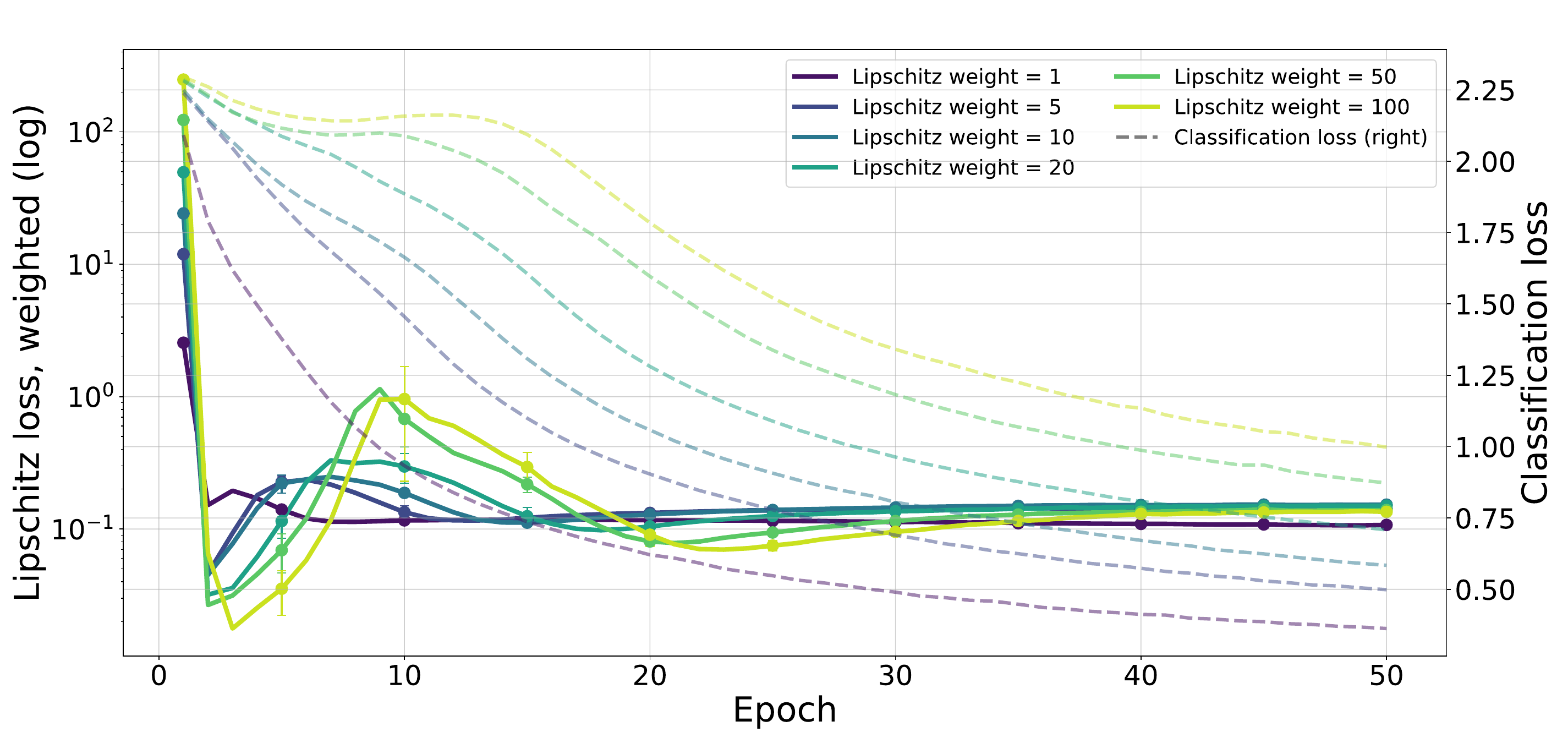}
		\caption{CIFAR-10}\label{fig:comparison_w_cifar}
	\end{subfigure}
	\\[1ex]
	\begin{subfigure}[b]{0.48\textwidth}
		\includegraphics[width=\textwidth]{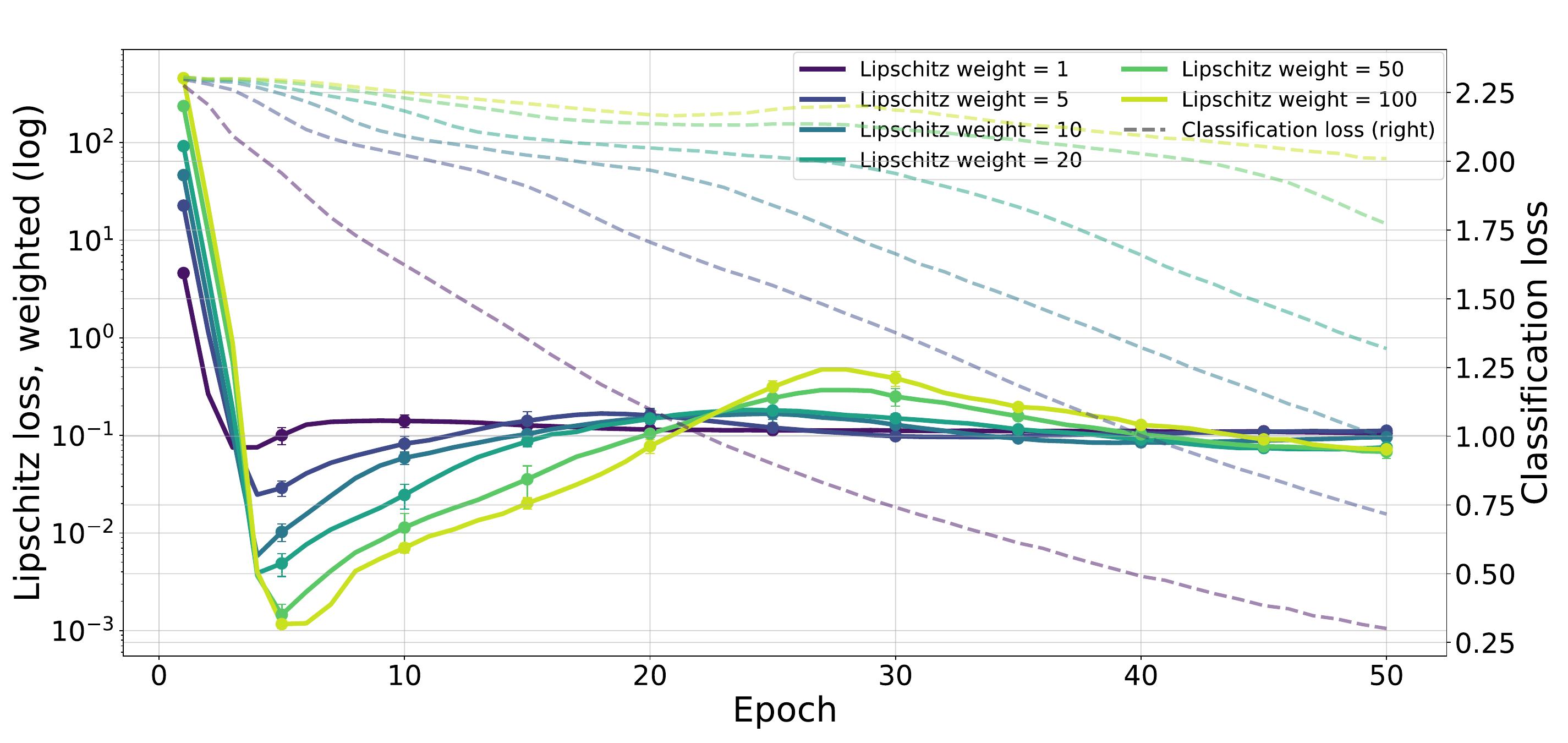}
		\caption{Imagenette-160}\label{fig:comparison_w_imnet}
	\end{subfigure}
	\caption{Effect of Lipschitz weight $\lambda \in \{1,5,10,20,50,100\}$ at fixed $L = 0.032$, over 50 epochs. Solid curves: weighted Lipschitz loss (left, log axis). Dashed: classification loss (right axis).}
	\label{fig:comparison_w}
\end{figure}

\begin{figure}
	\centering
	\begin{subfigure}[b]{0.48\textwidth}
		\includegraphics[width=\textwidth]{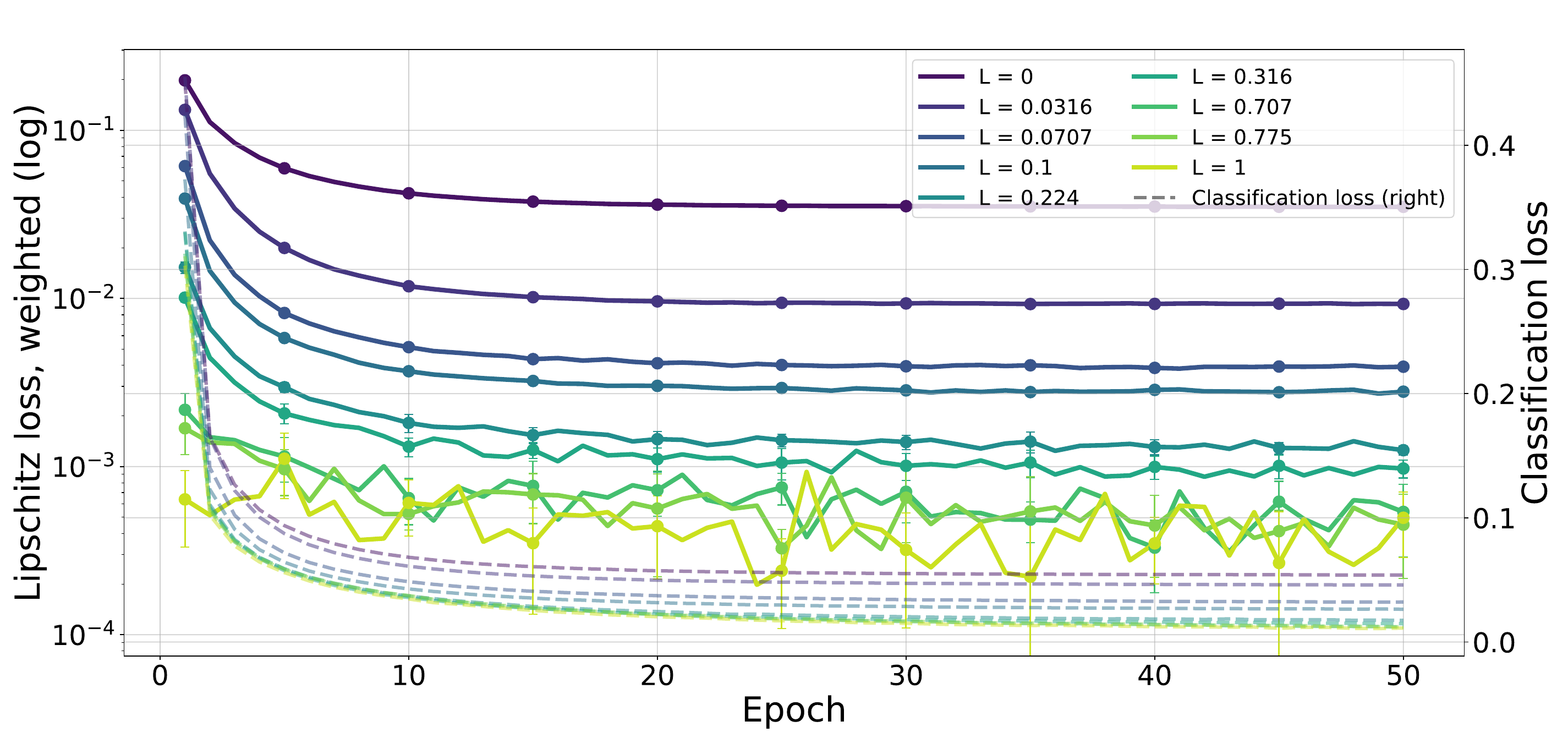}
		\caption{MNIST}\label{fig:comparison_l_mnist}
	\end{subfigure}
	\hfill
	\begin{subfigure}[b]{0.48\textwidth}
		\includegraphics[width=\textwidth]{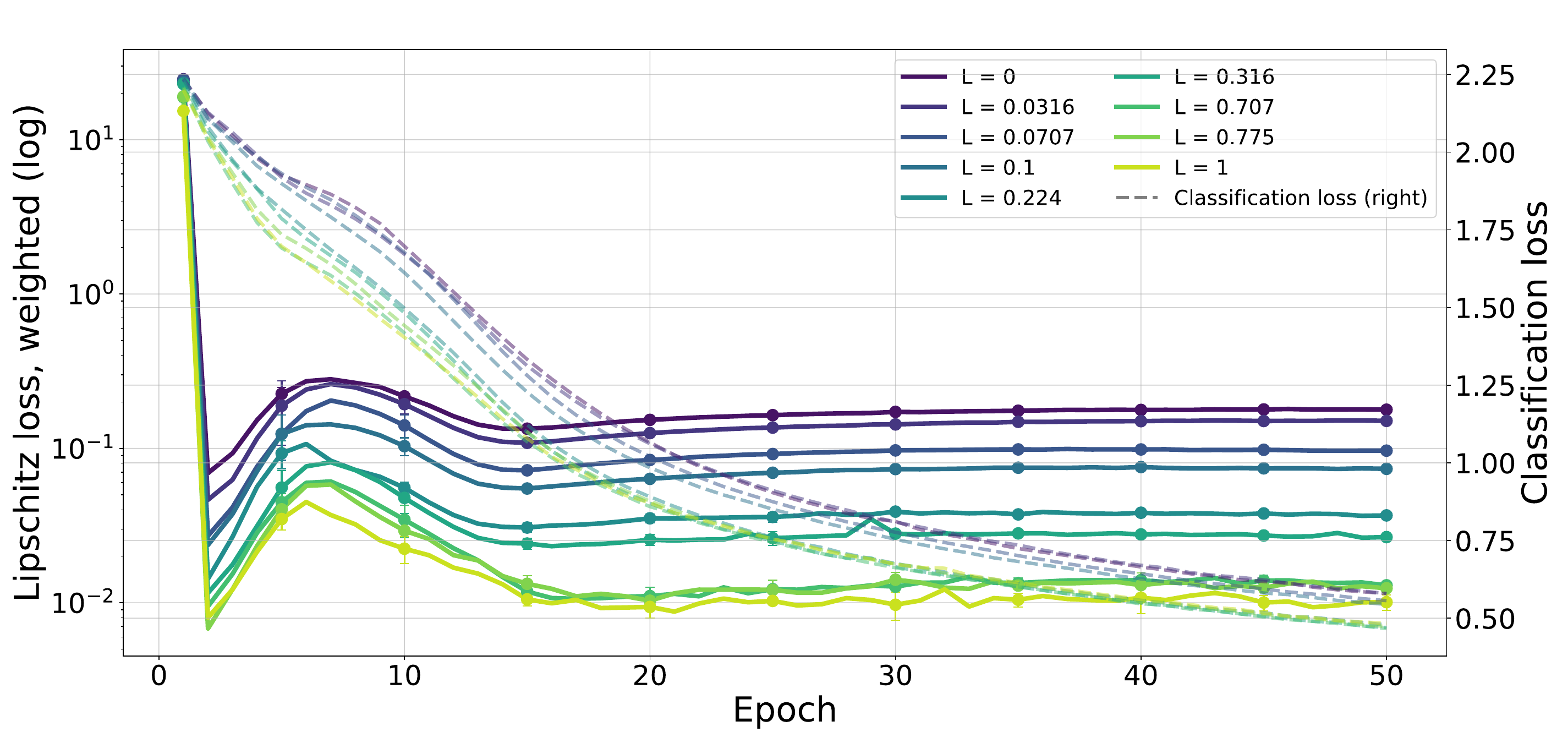}
		\caption{CIFAR-10}\label{fig:comparison_l_cifar}
	\end{subfigure}
	\\[1ex]
	\begin{subfigure}[b]{0.48\textwidth}
		\includegraphics[width=\textwidth]{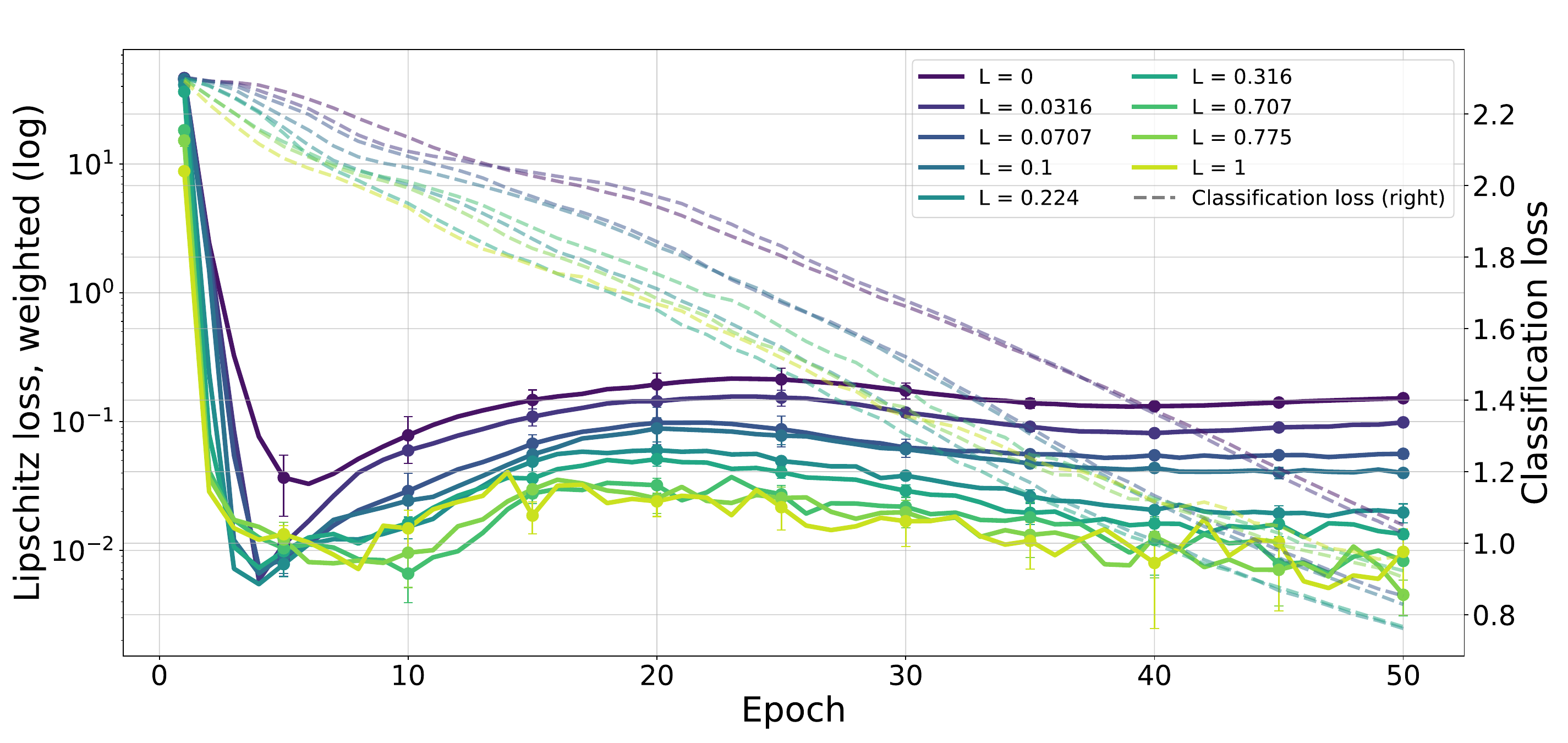}
		\caption{Imagenette-160}\label{fig:comparison_l_imnet}
	\end{subfigure}
	\caption{Effect of target Lipschitz constant $L \in \{0, 0.0316, 0.0707, 0.1, 0.224, 0.316, 0.707, 0.775, 1\}$ at fixed $\lambda$. Larger $L$ leaves the hinge slack and the Lipschitz loss inactive.}
	\label{fig:comparison_l}
\end{figure}

\begin{figure}
	\centering
	\begin{subfigure}[b]{0.48\textwidth}
		\includegraphics[width=\textwidth]{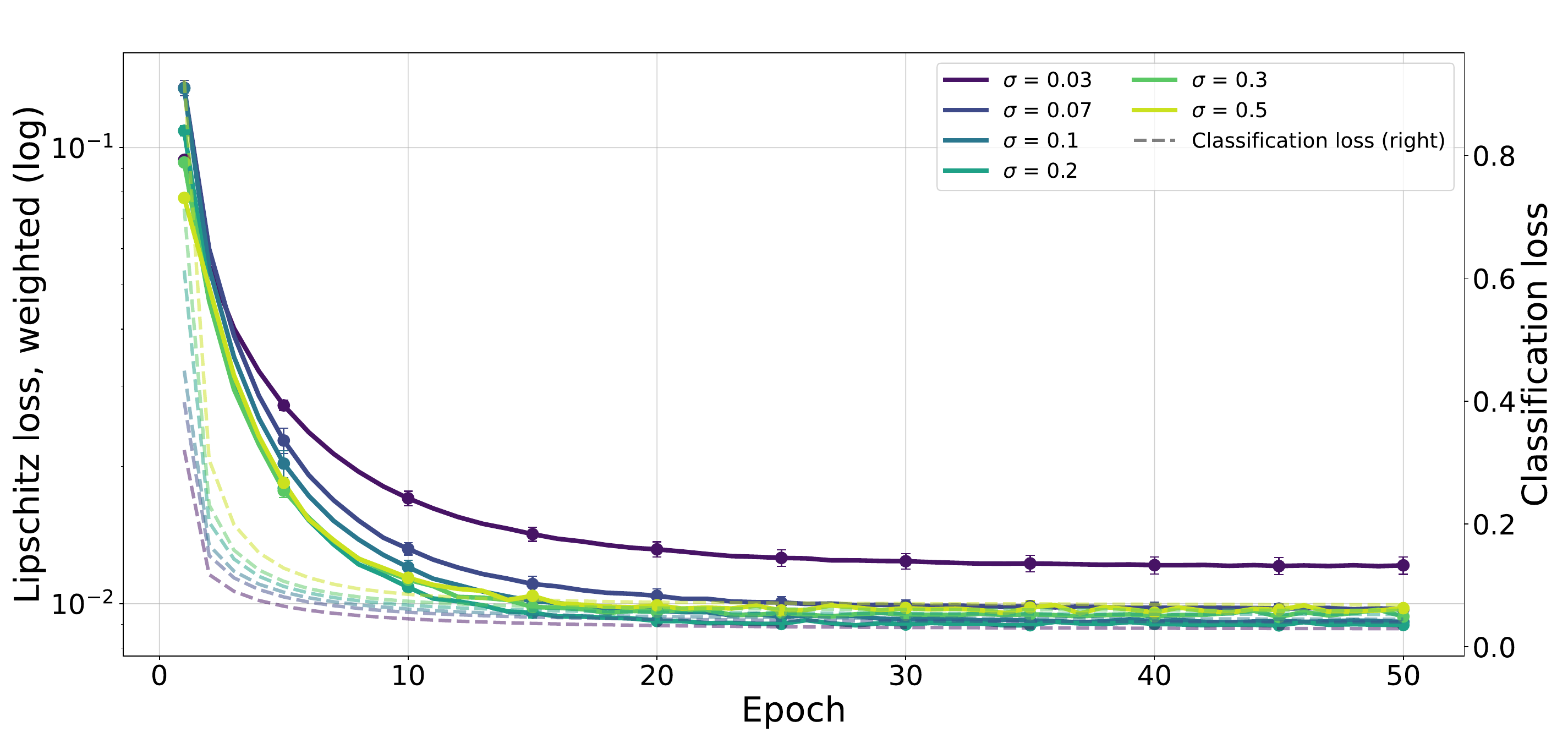}
		\caption{MNIST}\label{fig:comparison_sigma_mnist}
	\end{subfigure}
	\hfill
	\begin{subfigure}[b]{0.48\textwidth}
		\includegraphics[width=\textwidth]{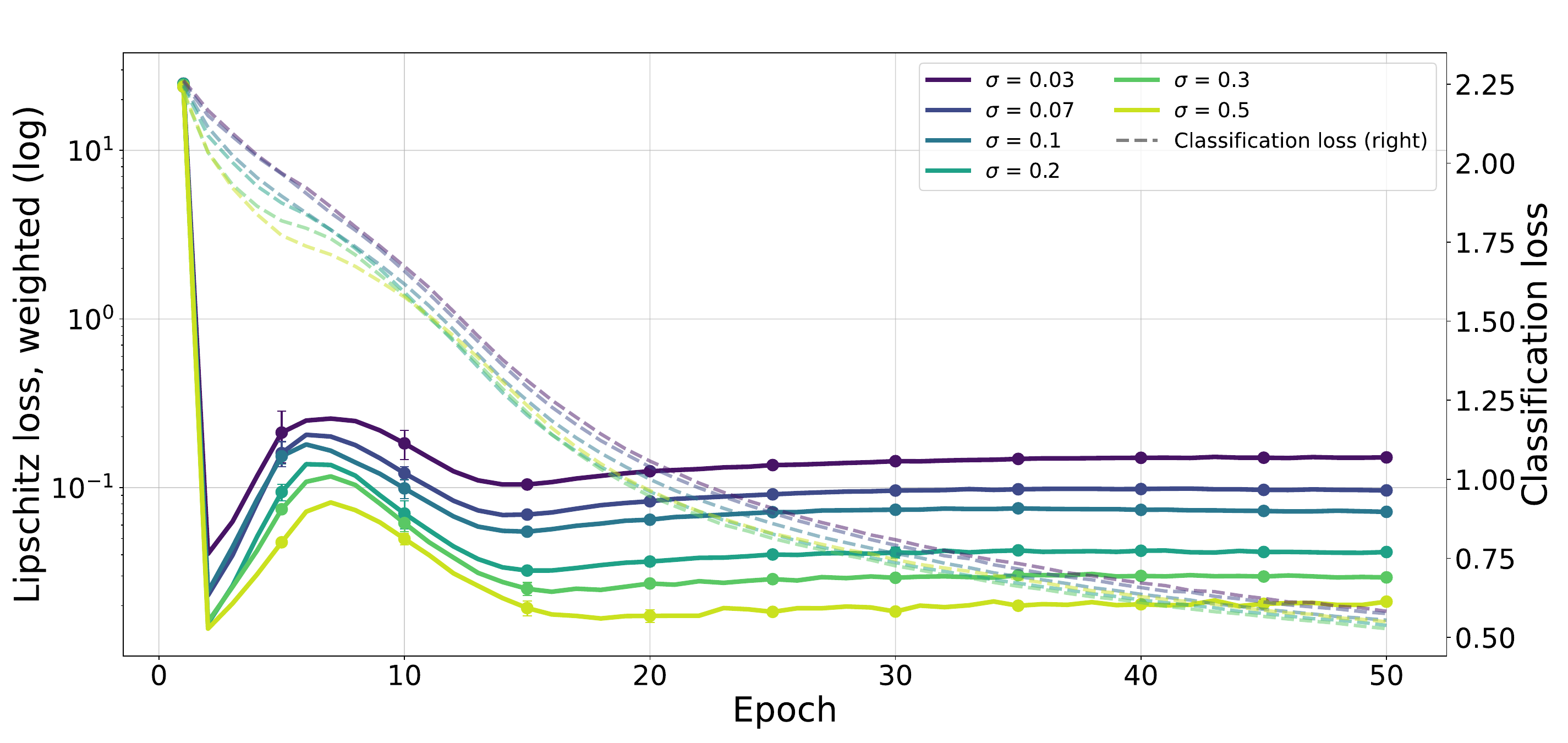}
		\caption{CIFAR-10}\label{fig:comparison_sigma_cifar}
	\end{subfigure}
	\\[1ex]
	\begin{subfigure}[b]{0.48\textwidth}
		\includegraphics[width=\textwidth]{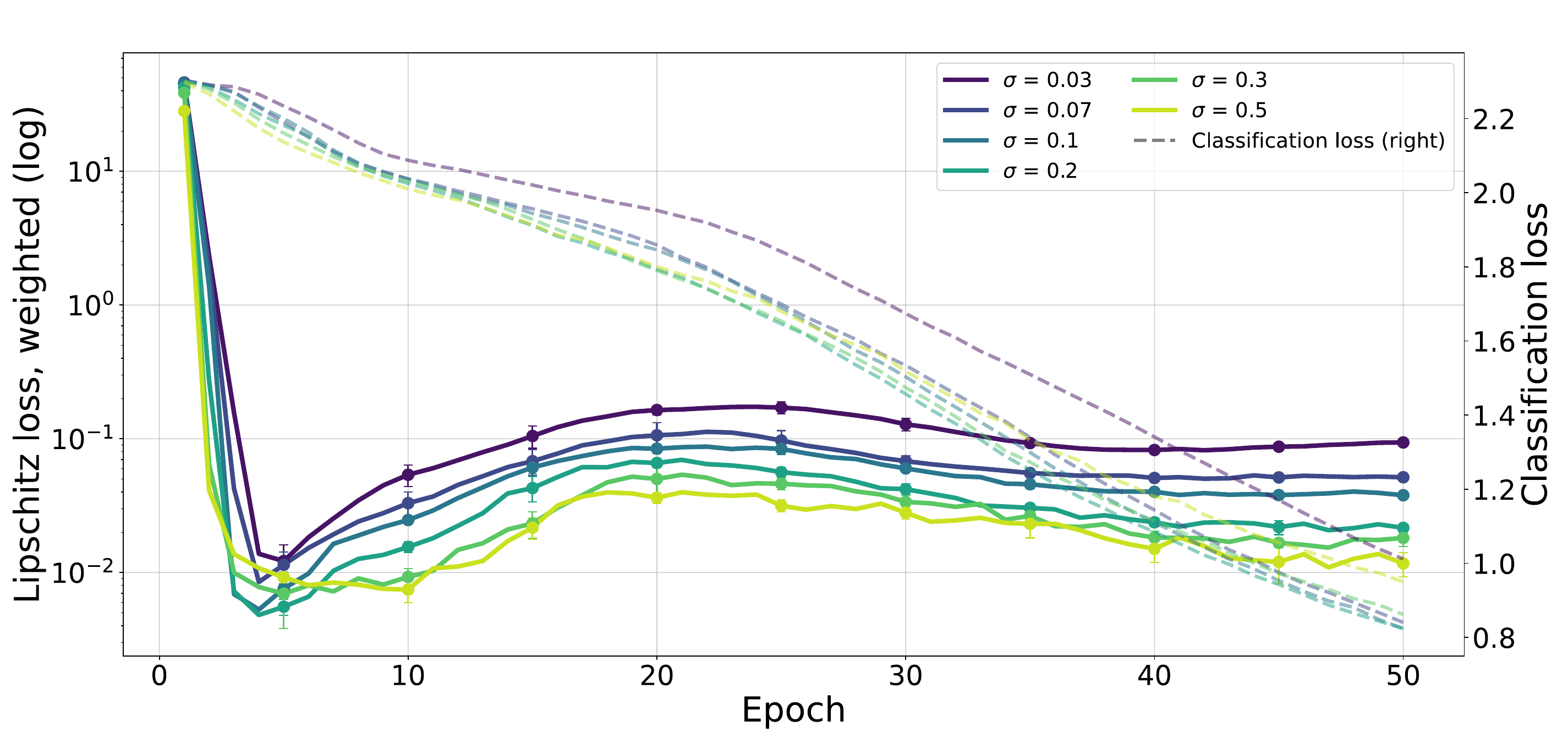}
		\caption{Imagenette-160}\label{fig:comparison_sigma_imnet}
	\end{subfigure}
	\caption{Effect of the neighborhood standard deviation
	$\sigma \in \{0.03, 0.07, 0.1, 0.2, 0.3, 0.5\}$ at fixed $L = 0.032$ and
	$\lambda = 10$, over 50 epochs. Solid curves: weighted Lipschitz loss
	$\lambda \cdot \mathcal{L}_\text{Lipschitz}$ (left, log axis). Dashed:
	classification loss (right axis). Mean over 5 folds; error bars give one
	standard deviation across folds.}
	\label{fig:comparison_sigma}
\end{figure}

The Lipschitz loss depends on three free parameters: the desired Lipschitz constant $L$, the weight $\lambda$, and the neighborhood standard deviation $\sigma$. Because these interact in a way that is not obvious from the loss definition alone, we visualize their effect in 
Figures~\ref{fig:comparison_w}, \ref{fig:comparison_l} and~\ref{fig:comparison_sigma}.

\textbf{Lipschitz weight $\lambda$.} On MNIST (Fig.~\ref{fig:comparison_w_mnist}) the weighted Lipschitz loss rises over the first 5--10 epochs, peaks, and then decays. This is the signature of the two objectives competing. At initialization, the embedding network is near-random, the embedding is compressed, and the constraint is barely violated; as the classification loss falls, the embedding network expands the embedding to separate classes, increasing its local sensitivity. The Lipschitz loss only begins to dominate once this expansion has occurred. The curves have the same shape for all $\lambda$ values; only their magnitude changes. Thus, $\lambda$ controls the strength of the Lipschitz loss but does not change how the loss behaves over time. The classification-loss curves for all six $\lambda$ are indistinguishable and converge to near zero by epoch 30, meaning that, for MNIST, $\lambda$ can be increased without a strong impact on the final training loss. Accordingly, the test accuracy also hardly changes across different $\lambda$ values (Table~\ref{tab:clean}).

\textbf{Lipschitz constant $L$:} Figure~\ref{fig:comparison_l} shows a monotone ordering across all three datasets: the smaller the Lipschitz constant $L$, the larger the residual Lipschitz loss at convergence.

For $L \gtrsim 0.707$, the constraint has almost no effect, indicating that the regular training of $\phi$ already ensures that distances are never enlarged beyond this factor. For example, on MNIST (Fig.~\ref{fig:comparison_l_mnist}), when $L$ is large ($0.707$, $0.775$, or $1$), the Lipschitz loss drops below $10^{-4}$ and often drops to zero and becomes unstable, with error bars varying across multiple orders of magnitude. Correspondingly, the constraint is violated on $25\%$ of sampled pairs on MNIST and $95\%$ on CIFAR-10 at our operating point (Table~\ref{tab:violation}). In practice, the Lipschitz loss effectively vanishes, so it contributes almost no gradient and we would not expect the learned embedding to differ from the unconstrained one. We did not test this directly at these values of $L$.

At the other extreme, the $L = 0$ and $L = 0.032$ curves are nearly identical on all three datasets. This is worth noting because $L = 0$ punishes any distance between $\phi(x)$ and $\phi(y)$, boiling down to something akin to L2 regularization on $\phi$. Still, it appears that the constraint degrades gracefully toward $L=0$, indicating that the models do not collapse. Clean accuracy stays high, and on MNIST the classification loss is unaffected. Likely, this stability can be effect that the classification loss and the Lipschitz loss act as opposing forces: collapsing all distances would lead to high classification loss, such that a stable tradeoff occurs. Where the tradeoff lies is, in turn, regulated by $\lambda$.

Two considerations govern the choice of $L$. Theorem~\ref{thm:main} guarantees local $\epsilon/L$-robustness, so we want $L$ to be as small as possible to achieve the highest possible robustness; at the same time, a smaller $L$ constrains the embedding network more strongly and, thus, may reduce accuracy. Figure~\ref{fig:comparison_l} shows how the Lipschitz loss develops for different $L$. For $L \gtrsim 0.707$, the Lipschitz loss collapses below $10^{-4}$ and becomes erratic, so the constraint is essentially never violated and contributes almost no gradient. At the other end, the $L = 0$ and $L = 0.032$ curves nearly coincide, indicating that the Lipschitz loss saturates below $0.032$ and reducing $L$ further changes nothing. Within the active range, the dashed classification-loss curves are largely unaffected by $L$, and clean accuracy stays within one percentage point of the baseline (Table~\ref{tab:clean}). We therefore adopt $L = 0.032$: the smallest value at which the Lipschitz loss is still distinguishable from the degenerate $L = 0$ case and at which accuracy is still little affected.

Note that $L$ is \emph{not} the actual Lipschitz constant of the trained embedding network. $L$ would only be the Lipschitz constant if the Lipschitz loss would reach zero for infinitely long training (i.e.: infinitely many stochastic samples across the neighborhood). Figure~\ref{fig:comparison_l} shows a non-zero residual for every $L \leq 0.224$. The constraint is violated on $25\%$ of sampled pairs on MNIST and $95\%$ on CIFAR-10 (Table~\ref{tab:violation}). To quantify how far the trained network is from the requested bound, we report the empirical stretch ratio $\lVert\phi(x)-\phi(y)\rVert / \lVert x-y\rVert$, averaged over exactly those Gaussian neighbor pairs $(x, y)$ on which the Lipschitz loss is evaluated. We emphasize that this is \emph{not} the Lipschitz constant of $\phi$, which is NP-hard to compute even for two-layer networks~\cite{scaman2018lipschitz}. It is a sample average over a finite set of directions and therefore a lower bound on the true constant (the maximum over all pairs can only be larger). Its mean value is $0.028$ on MNIST and $0.081$ on CIFAR-10, against $L = 0.032$ -- close on MNIST, exceeded by a factor of $2.5$ on CIFAR-10. Consequently, $\epsilon/L$ is not a certified radius, and the robustness reported in Section~\ref{sec:results} is empirical throughout.  This matters because $L \to 0$ is the degenerate limit where all points are mapped to the same constant embedding~\cite{zuhlke2025adversarial}.
\begin{table}[ht]
  \centering
\caption{How strictly the Lipschitz constraint is enforced as the weight $\lambda$ varies, at fixed $L = 0.032$. Two quantities measure this on the same sampled training pairs: ``violation'' counts the fraction of pairs for which the constraint of Eq.~\eqref{eq:lipschitz_loss} is not satisfied, while ``mean stretch ratio'' reports by how much the embedding is stretched on those pairs. Values are taken at the final training epoch and averaged over 5 folds.}
  \label{tab:violation}
  \resizebox{\linewidth}{!}{%
    \begin{tabular}{lcccc}
      \toprule
      Dataset & $\lambda$ & Violation & Mean stretch ratio & Clean accuracy (\%) \\
      \midrule
      MNIST
        & $1$   & $0.867$ & $0.049$ & $98.47$ \\
        & $5$   & $0.410$ & $0.032$ & $98.33$ \\
        & $10$  & $0.252$ & $0.028$ & $98.32$ \\
        & $20$  & $0.147$ & $0.025$ & $98.13$ \\
        & $50$  & $0.070$ & $0.023$ & $98.21$ \\
        & $100$ & $0.041$ & $0.021$ & $98.08$ \\
      \midrule
      CIFAR-10
        & $1$   & $1.000$ & $0.200$ & $78.12$ \\
        & $5$   & $0.988$ & $0.107$ & $77.17$ \\
        & $10$  & $0.952$ & $0.081$ & $75.54$ \\
        & $20$  & $0.839$ & $0.061$ & $73.56$ \\
        & $50$  & $0.590$ & $0.043$ & $70.17$ \\
        & $100$ & $0.401$ & $0.034$ & $67.00$ \\
      \midrule
      Imagenette-160
        & $1$   & $0.983$ & $0.103$ & $76.36$ \\
        & $5$   & $0.770$ & $0.054$ & $72.31$ \\
        & $10$  & $0.551$ & $0.041$ & $66.49$ \\
        & $20$  & $0.334$ & $0.031$ & $56.11$ \\
        & $50$  & $0.189$ & $0.026$ & $40.09$ \\
        & $100$ & $0.131$ & $0.023$ & $30.69$ \\
      \bottomrule
    \end{tabular}%
  }
\end{table}

Table~\ref{tab:violation} shows that the violation rate (measured on randomly sampled Gaussian neighbors of the training points, using the same pairs the Lipschitz loss is computed on) falls monotonically with $\lambda$ on all three datasets. This confirms that $\lambda$ controls how strictly the constraint is enforced, and suggests that the violation rate is never driven to zero, so target $L$ is not attained everywhere in either setting. The cost differs sharply by dataset: raising $\lambda$ from $1$ to $100$ reduces clean accuracy by $0.4$ points on MNIST, $11$ on CIFAR-10, and $46$ on Imagenette-160, which is why we adopt $\lambda = 10$ rather than a larger value. 

\textbf{Neighborhood standard deviation $\sigma$.} The third parameter controls where the constraint is sampled rather than how tightly it is enforced. If $\sigma$ is increased, neighbors $y_i$ are drawn further away from the training point $x_i$, so the Lipschitz condition of Eq.~\eqref{eq:lipschitz_loss} is checked over a wider region. Figure~\ref{fig:comparison_sigma} shows a monotone ordering on CIFAR-10 and Imagenette-160. The results indicate that the larger $\sigma$, the smaller the residual Lipschitz loss at convergence, spanning about one order of magnitude between $\sigma = 0.03$ and $\sigma = 0.5$. This is a property of the loss rather than of the embedding network, because the term $L^2 d(x_i, y_i)^2$ grows quadratically with the distance to the neighbor while the embedding displacement saturates. Consequently, the constraint is satisfied more often for distant neighbors, and the Lipschitz loss in Equation~\eqref{eq:lipschitz_loss} is inactive on those neighbor pairs. A small residual at large $\sigma$ therefore indicates that the constraint is loosely enforced. On MNIST, only $\sigma = 0.03$ is separated from the rest, and the remaining values coincide, indicating that the Lipschitz loss is insensitive to $\sigma$ above $0.1$ on this dataset. The classification loss behaves in the opposite direction on CIFAR-10 and Imagenette-160. On both of these datasets, the smallest $\sigma$ converges most slowly, and the ordering is preserved throughout training, while on MNIST the curves are indistinguishable. We therefore select $\sigma = 0.1$ for MNIST and $\sigma = 0.03$ for CIFAR-10 and Imagenette-160, which are the smallest values at which the Lipschitz loss is still clearly active.

\section{Discussion}\label{sec:discussion}

The results indicate that adversarial robustness is not the consequence of a single design choice or hyperparameter. Several factors act together, and in our study the classification head emerges as the dominant one, rather than the Lipschitz loss alone. Under FGSM and PGD attacks at moderate and large budgets, the two prototype-based models (GLVQ and Lipschitz GLVQ) are more robust than the linear and SVM classifiers trained on the same frozen embedding. Standard GLVQ, without any Lipschitz loss, already outperforms both Linear and SVM under FGSM, PGD, and C\&W attacks. This reproduces the observation by Saralajew et al.~\cite{SaralajewGLVQ} that GLVQ models possess a high base robustness before any explicit certification, and it is consistent with the margin-based view of these models: the decision rule of a nearest prototype classifier carries its own hypothesis margin, whereas linear+softmax clasification carries none~\cite{crammer2002, hammer2005}. This gain is obtained at a cost to clean accuracy, which remains below two percentage points on all three benchmarks (Table~\ref{tab:clean}), and without giving up interpretability, since the prototypes are stored in $\mathcal{X}$ and are initialized from training images. 
The Lipschitz constraint plays a complementary role: it is what makes the guarantee of Theorem~\ref{thm:main} available at all. Without a controlled value of $L$, that is, without a bound on how much the embedding network can stretch distances in the input space, robustness guarantees in the embedding space tell us nothing about robustness in the original input space, because an unconstrained embedding network can map an arbitrarily small input perturbation to an arbitrarily large attack in the embedding space, thus exceeding the robustness guarantees. 
The measured effect of the Lipschitz loss on robust accuracy, however, depends strongly on the choice of classification head, even though the embedding itself is identical for all three. The improvement of Lipschitz-Linear and Lipschitz-SVM over their unconstrained counterparts is confined to MNIST. On CIFAR-10, Lipschitz-Linear and Lipschitz-SVM fail to outperform their non-Lipschitz counterparts, and on Imagenette-160 the constrained variants are marginally worse against stronger attacks. The two margin-based heads on top of the embedding model therefore do not benefit equally from the Lipschitz loss. 
Figure~\ref{fig:comparison_l} also visualizes the range in which the Lipschitz loss has any effect. For $L \gtrsim 0.707$, the loss falls below $10^{-4}$, indicating that the embedding network that does not stretch distances beyond that factor. So setting $L$ this large does not have any effect; the regular training of the embedding network would have ensured this Lipschitz constant anyway. By contrast, for $L = 0.032$, the Lipschitz loss remains nonzero throughout training on all three datasets. The difference between Lipschitz-GLVQ and standard GLVQ on MNIST and CIFAR-10 can thus be attributed to the effect of the Lipschitz loss. More broadly, Figure~\ref{fig:comparison_l} indicates that the Lipschitz loss has no effect for $L \geq 0.707$ and saturates for $L \leq 0.032$. Hence, we select $L = 0.032$. For this setting, the clean (i.e. not attacked) accuracy remains almost unaffected while achieving as strong robustness as possible. 
For example, on MNIST, where the input dimension is low and the embedding space is only $16$-dimensional, Lipschitz-GLVQ retains roughly $52.9\%$ accuracy under PGD attacks at $\epsilon = 0.20$ (Section~\ref{sec:results}), an attack strength at which every unconstrained model has collapsed to zero.

Computationally, Lipschitz training is relatively cheap: it costs one additional forward pass through $\phi$ per mini-batch, against the several forward--backward passes per sample required by the inner maximization of adversarial training~\cite{madry2017towards}. On Imagenette-160, this cheapness is partly offset by slower convergence, as the constrained embedding network reaches only $65.1\%$ validation accuracy at 50 epochs, compared to $74.7\%$ for the baseline.

As discussed in Section~\ref{sec:hyperparams_selection}, the Lipschitz constant $L$, the Lipschitz weight $\lambda$, and the neighborhood width $\sigma$ do not act independently, making hyperparameter optimization a multi-dimensional (and multi-objective; due to the accuracy-robustness tradeoff) optimization problem. Still, it is possible to characterize the different roles of $L$, $\lambda$, and $\sigma$, and provide some guidance for reasonable settings. $L$ limits how much the embedding network may stretch a small input perturbation into a displacement in the embedding space, $\lambda$ determines how aggressively that limit is enforced during training, and $\sigma$ decides whether the constraint is checked only very close to the training data or over a broader region. A larger $\lambda$ enforces the constraint more strictly and drives the violation rate down, while a larger $L$ relaxes it, so the two can effectively cancel. If $\sigma$ is too small, the embedding network may satisfy the Lipschitz bound 
near the training points while remaining unconstrained further away, leaving gaps that an attack can exploit. A small $L$ therefore does not guarantee robustness if $\lambda$ is too weak to enforce it or $\sigma$ is too narrow to cover the relevant region, just as a large $\lambda$ with a wide $\sigma$ is of little use if $L$ is set so loosely that the constraint is vacuous. Only when all three are chosen in concert does the constraint provide a transferable statement about robustness. The three parameters are not tunable independently. Reporting only one without the other two is insufficient for reproducibility. Two additional observations follow. First, the architecture of the embedding network may be just as important as the Lipschitz loss.  We used a small embedding network for fair comparison, not for peak performance. A larger embedding network may raise both clean and robust accuracy. On MNIST, clean accuracy is high ($98\%$), so the model handles a wide range of attacks. On CIFAR-10 and Imagenette-160, clean accuracy is lower (around $70\%$). There is less room to spare, so performance drops faster. Second, the interaction differs by dataset, so a single global setting of $(L, \lambda, \sigma)$ is unlikely to be appropriate everywhere.

The behavior across datasets and attacks is not uniform. Lipschitz-GLVQ is the strongest model under FGSM attacks on all three datasets once the attack limit exceeds a small threshold. Under PGD attacks, the six models are close on MNIST at small budgets but separate sharply at larger ones, while on CIFAR-10 Lipschitz-GLVQ retains a clear advantage over the linear and SVM heads. On Imagenette-160, however, Lipschitz-SVM is the strongest model under PGD attacks over the range where accuracies remain substantial, with Lipschitz-GLVQ second. Since both are maximum-margin heads, which Theorem~\ref{thm:main} treats alike, we take this as another instance of the SVM/GLVQ difference that needs to be explored further. The C\&W results are the least consistent: on MNIST the ordering is clear, with the mean $\ell_2$ distortion of the GLVQ pair rising from $1.57$ to $2.93$, and on Imagenette-160 it is favorable ($0.387 \rightarrow 0.414$), whereas on CIFAR-10 the two are indistinguishable ($0.194$ against $0.194$). 
The C\&W is also the least stable attack across runs. Between two runs on Imagenette-160, the \emph{unconstrained} baselines shifted as well (Linear $0.36 \rightarrow 0.46$, SVM $0.32 \rightarrow 0.40$), although they contain no Lipschitz loss and are unaffected by $L$ and $\lambda$. Part of the variation therefore originates in the internal routines of the C\&W attacks, rather than in the models (Section~\ref{sec:comparison}). Beyond this, $\sigma$ is expressed in fixed input units and so covers a smaller \emph{relative} region of a $64\times64\times3$ image than of a $28\times28$ digit.
Resolution-aware choice of $\sigma$ would be a natural next step. Imagenette-160 also has by far the smallest training set ($9,469$ images), and the matched 50-epoch budget penalizes the constrained embedding network on this dataset given its slower convergence. The FGSM results on this dataset, which are stable across folds, show Lipschitz-GLVQ ahead of both linear heads, and we take those as the primary evidence.

Overall, our key results are: The combination of Lipschitz training for a deep embedding network and a GLVQ classifier on the embeddings yields a promising deep prototype classification approach. Our theory suggests that this is due to a transfer of robustness guarantees in the embedding space to the overall model, thanks to Lipschitz continuity. However, these results are complicated by the fact that Lipschitz-SVM falls short of Lipschitz-GLVQ (even though we would have expected similar results) and that GLVQ performs very similar to Lipschitz-GLVQ for some dataset-attack-combinations.

\subsection{Limitations}

While we hope to have substantially pushed forward the research on deep embedding networks combined with robust, shallow classifiers, we also acknowledge multiple limitations:

\begin{itemize}
\item Local robustness is assumed by our theory, not proven. We rely on prior robustness proofs. However, the fact that Lipschitz SVM does not perform as well as Lipschitz-GLVQ indicates that this may be problematic. A deeper theoretical analysis of the difference between SVM and GLVQ robustness guarantees is required.
\item Lipschitz training does not guarantee a local Lipschitz bound but only punishes violations for stochastically sampled neighbors. This leaves room for future work on more effective training schemes (without losing the computational efficiency of our approach).
\item While empirical improvements are striking against FGSM attacks, robustness gains against stronger attacks (PGD and C\&W) are more modest. Accordingly, stronger guarantees may be needed.
\item Our evaluation is limited to three attacks, each using a single random restart. A more comprehensive evaluation using an ensemble-based approach such as AutoAttack~\cite{croce2020reliable} would provide a stronger test of the observed robustness and is left for future work.
\item Our experiments are limited to three data sets and three classifiers. Further classifiers, especially those with robustness guarantees, could be investigated in subsequent work. Similarly, further datasets, involving more modalities than just image data, could be investigated.
\end{itemize}

\section{Conclusion}

Broadly speaking, our motivation stems from the idea that robustness and interpretability are connected by a similar underlying question: does a model only change its prediction when the input has accumulated enough evidence to support a different label? The MNIST results provide some support for this interpretation. In this setting, Lipschitz-GLVQ changes its prediction only at perturbation levels where the digit itself becomes difficult for a human to recognize. This correspondence is less clear for CIFAR-10 and Imagenette-160. For these datasets, a human observer would generally still assign the original label within the tested perturbation budgets, even though the models may already have changed their predictions. Therefore, while Lipschitz-GLVQ improves robustness on the color datasets, the improvement is not yet strong enough to fully support the proposed interpretability argument, and the intersection of robustness and interpretability remains a promising direction for future work.

\bibliographystyle{plainnat}
\bibliography{literature}
\end{document}